\documentclass[lettersize,journal]{IEEEtran}
\usepackage{amsmath,amsfonts}
\usepackage[linesnumbered,ruled,vlined]{algorithm2e}
\SetCommentSty{mycommfont}
\newcommand{\func}[2]{\textcolor{blue}{\texttt{#1}}(#2)}

\let\oldnl\nl
\newcommand{\nonl}{\renewcommand{\nl}{\let\nl\oldnl}}
\usepackage[font=footnotesize]{caption}
\usepackage{array}
\usepackage[caption=false,font=normalsize,labelfont=sf,textfont=sf]{subfig}
\usepackage{textcomp}
\usepackage{url}
\usepackage{verbatim}
\usepackage{enumitem}
\usepackage{graphicx}
\usepackage{cite}
\usepackage{placeins}
\usepackage[colorlinks=true, linkcolor=magenta, citecolor=magenta, urlcolor=magenta]{hyperref}
\usepackage{threeparttable}
\usepackage{longtable}
\usepackage{multirow}
\usepackage{float}
\usepackage{array}
\usepackage{booktabs}
\usepackage{tipa}
\usepackage{supertabular}
\usepackage[export]{adjustbox} 
\usepackage{booktabs}
\usepackage{setspace}
\usepackage{mathrsfs}
\usepackage{placeins}
\usepackage{amsmath}
\usepackage{array}
\usepackage{amssymb}
\usepackage{amsthm}
\usepackage{microtype}
\usepackage{url}
\usepackage{amsfonts,amssymb}
\usepackage{dsfont}
\usepackage{mathtools}
\usepackage{makecell}

\newcommand{\LaTeXstyle}[1]{{\textsc{#1}}}
\usepackage{listings}
\usepackage{multirow}
\usepackage{dblfloatfix}
\usepackage[table,xcdraw]{xcolor}
\usepackage{colortbl}
\usepackage{array}

\usepackage{tgheros} 

\begin{document}

\providecommand{\SetAlgoCommentTemplate}[1]{}

\definecolor{codebg}{RGB}{245, 245, 245} 
\definecolor{keyword}{RGB}{204,104,67} 
\definecolor{comment}{RGB}{153, 153, 153} 
\definecolor{string}{RGB}{107,152,184} 
\definecolor{number}{RGB}{139, 69, 19} 
\definecolor{methodname}{RGB}{255, 165, 0} 
\definecolor{output}{RGB}{255, 0, 0} 
\definecolor{color1}{gray}{0.0}   
\definecolor{color2}{gray}{0.1}
\definecolor{color3}{gray}{0.2}
\definecolor{color4}{gray}{0.3}
\definecolor{color5}{gray}{0.4}
\definecolor{color6}{gray}{0.5}
\definecolor{color7}{gray}{0.6}
\definecolor{color8}{gray}{0.7}
\definecolor{color9}{gray}{0.75}
\definecolor{color10}{gray}{0.8}
\definecolor{color11}{gray}{0.85}
\definecolor{color12}{gray}{0.9}  

\title{Enhancing LLMs for Power System Simulations: \\ A Feedback-driven Multi-agent Framework}

\author{Mengshuo~Jia,~\IEEEmembership{Member,~IEEE},
Zeyu~Cui,~\IEEEmembership{Member,~IEEE},~and
Gabriela~Hug,~\IEEEmembership{Senior~Member,~IEEE}





}

\markboth{Journal of \LaTeX\ Class Files, May~2025}%
{Shell \MakeLowercase{\textit{et al.}}: A Sample Article Using IEEEtran.cls for IEEE Journals}


\maketitle

\begin{abstract}

The integration of experimental technologies with large language models (LLMs) is transforming scientific research. It positions AI as a versatile research assistant rather than a mere problem-solving tool. In the field of power systems, however, managing simulations --- one of the essential experimental technologies --- remains a challenge for LLMs due to their limited domain-specific knowledge, restricted reasoning capabilities, and imprecise handling of simulation parameters. To address these limitations, this paper proposes a feedback-driven, multi-agent framework. It incorporates three proposed modules: an enhanced retrieval-augmented generation (RAG) module, an improved reasoning module, and a dynamic environmental acting module with an error-feedback mechanism. Validated on 69 diverse tasks from \LaTeXstyle{Daline} and \LaTeXstyle{MATPOWER}, this framework achieves success rates of 93.13\% and 96.85\%, respectively. It significantly outperforms ChatGPT 4o, o1-preview, and the fine-tuned GPT4o, which all achieved a success rate lower than 30\% on complex tasks. Additionally, the proposed framework also supports rapid, cost-effective task execution, completing each simulation in approximately 30 seconds at an average cost of 0.014 USD for tokens. Overall, this adaptable framework lays a foundation for developing intelligent LLM-based assistants for human researchers, facilitating power system research and beyond.

\end{abstract}

\IEEEpeerreviewmaketitle

\section{Introduction}\label{sec:Intro}
\IEEEPARstart{C}{ombining} laboratory automation technologies with large language models (LLMs) enables automated execution of scientific experiments \cite{boiko2023autonomous}. Related advances span the fields of mathematics, chemistry, and clinical research, including mathematical algorithm evolution \cite{romera2024mathematical}, geometry theorem proving \cite{trinh2024solving}, chemical experiment design and execution \cite{boiko2023autonomous}, as well as the development and validation of machine learning approaches for clinical studies \cite{tayebi2024large}. These recent achievements signal a new research paradigm. {\color{black}That is}, positioning AI as a research assistant for humans with natural language communication abilities, rather than merely a specialized problem solver as in the past. Establishing LLMs as research assistants also holds significant potential for advancing power systems research.

Power systems research heavily relies on simulations. To develop LLM-based research assistants in this field, LLMs must be equipped with the capability to conduct power system simulations. Enabling LLMs to execute simulation tasks has multiple implications: (i) At the assistant level, LLMs capable of conducting simulations would allow researchers to focus more on idea-intensive activities, such as simulation design, rather than on labor-intensive tasks like simulation implementation. (ii) At the interface level, LLMs conducting simulations might offer a natural-language interface. This interface can connect simulation tasks with other upstream/downstream power system tasks using natural language as the input/output. This is particularly helpful when the original inputs and outputs of these tasks are heterogeneous (e.g., different modalities), which are originally challenging to program cohesively using regular codes. (iii) At the coding level, LLMs executing simulations might be a step toward natural language coding in power systems. This might signify an evolution in programming, bringing it closer to a more intuitive, language-driven approach, a long-standing goal of programming development for decades.

However, LLMs inherently lack the capability to perform power system simulations. For recently developed simulation tools not included in LLM pre-training datasets, LLMs generally cannot execute these simulations accurately. Even for well-established tools included in pre-training data, simulation precision remains unsatisfactory. For instance, GPT-4 often has difficulty creating small distribution grids using OpenDSS \cite{bonadia2023potential} or writing code for simple (optimal) power flow problems \cite{dong2024exploring}, even though information about both OpenDSS and (optimal) power flow is available within GPT-4's pre-training dataset. While the underlying causes of this issue have not been widely discussed and recognized in the energy domain, {\color{black}the following factors have been proposed} as potential explanations: 
\begin{itemize}[leftmargin=10pt]
    \item \textit{Frequency}: The low frequency of domain-specific power system knowledge in LLM training datasets --- especially in the long tail of rarely encountered data --- limits the models' ability to generalize effectively for specialized simulation tasks \cite{chang2024large}.
    \item \textit{Quality}: High-quality, instruction-tuned, or query-based coding data specific to power system simulations in available open-source data is lacking. Missing explanatory code annotations make it difficult for LLMs to fully contextualize and operationalize power system simulations. 
    \item \textit{Complexity}: {\color{black} Complex power system simulations require multi-step reasoning, which is inherently challenging. This difficulty increases when the model's learned patterns contain sparse or ambiguous representations related to power system simulations. }
    \item \textit{Precision}: Precise identification of simulation parameters, functions, and their logical connections, poses high demands on LLMs, especially when LLMs' knowledge about simulations is incomplete or fragmented. This may result in a semantic drift, causing LLMs' code generation to gradually deviate from the accurate version.
\end{itemize}

The challenges outlined above can be grouped into three main limitations: (i) limited simulation-specific knowledge, (ii) restricted reasoning capabilities for simulation tasks, and (iii) imprecision in function and option application. Enhancing the simulation capability of LLMs requires addressing these limitations. However, \textit{few existing works have explicitly focused on overcoming the above barriers to improve the simulation capability of LLMs}, even though LLM applications in power systems are growing rapidly. Specifically, in power systems,  LLMs have been recently used to translate language-based rules into mathematical constraints to facilitate optimal power flow (OPF) analysis, bridging the gap between rule-based and computational methods \cite{yan2023real}. Researchers have also leveraged LLMs to interpret decision-making processes in real-time {\color{black}markets}, enhancing the transparency of deep reinforcement learning systems by revealing decision rationales \cite{zhang2024large}. Additionally, LLMs have been employed to retrieve and summarize documents in response to specific power system queries \cite{huang2024large}. In other work, LLMs have been applied to derive OPF solutions iteratively by utilizing historical cost-solution data \cite{huang2024large}. LLMs have also facilitated gathering user preference w.r.t. electric {\color{black}vehicle} charging, where user inputs are integrated to refine functions for EV charging optimization problems \cite{huang2024large}. Furthermore, by combining LLMs with retrieval-augmented generation (RAG), researchers develop a carbon footprint accounting system capable of dynamically retrieving and integrating real-time, domain-specific carbon data \cite{wang2024carbon}. Moreover, for cybersecurity applications, LLMs have played a role in anomaly detection to enhance system security \cite{zaboli2024chatgpt}. In forecasting, LLMs, such as LLaMa2, have been used to integrate social event data into time series models, enhancing the accuracy and contextual relevance of predictions, for example, in electricity demand \cite{wang2024news}. Also, LLMs have supported perception analysis by evaluating media sentiment and public acceptance levels of solar power initiatives, providing insights into public opinion trends \cite{nuortimo2024global}. Moreover, a comprehensive benchmarking framework for LLMs has been proposed for the energy domain in \cite{zhou2024elecbench}, helping to establish standardized evaluation criteria for LLMs in energy-related applications. On the other hand, potential cybersecurity threats, arising due to the application of LLMs in power systems, have also been analyzed and summarized \cite{ruan2024applying}.  

\begin{figure}[t!]
\centering
\includegraphics[width=1\linewidth]{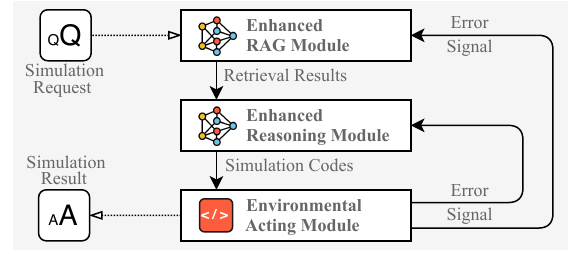} 
\caption{The feedback-driven multi-agent framework. It consists of an enhanced RAG module, an advanced reasoning module, and an {\color{black}environmental acting module}, all interconnected through an {\color{black}error-feedback} mechanism. This framework enables iterative refinement by incorporating simulation-specific knowledge, improving reasoning for complex simulation tasks, and facilitating environmental interaction to generate accurate simulation results.}
\label{fig:wholeframework}
\end{figure}

Despite the various applications mentioned above, only a few studies have focused directly on using LLMs for power system simulations. These studies, however, only focus on conceptualizing the potential of LLMs in the simulation field \cite{ding2023exploration}, showcasing their current capabilities \cite{huang2023large, ding2023exploration}, and assessing their effectiveness in generating general-purpose code for power system studies \cite{bonadia2023potential, dong2024exploring}. While these studies offer valuable insights, they did not address the limitations pointed out earlier that limit LLMs' simulation performance. Although the standard RAG approach used in \cite{RAG, ding2023exploration, dong2024exploring} can indeed enable LLMs to incorporate external power systems knowledge, it is unsuitable for simulation tasks. This happens because the standard RAG retrieves information based on the entire request as a single unit. As a result, it fails to capture the nuanced structure of complex simulation requests. It often conflates distinct function-related and option-related elements, leading to inefficiencies and reduced retrieval accuracy. Consequently, existing works fall short of systematically developing and advancing LLMs'  capability to handle complex power system simulations.

To bridge this gap and enhance LLMs' capability in power system simulations, this paper proposes a modular, feedback-driven, multi-agent framework. It integrates several innovative strategies, as shown in Fig.~\ref{fig:wholeframework}. Accordingly, this paper contributes in the following ways: 
\begin{itemize}[leftmargin=10pt]
    \item {\color{black}It proposes} an enhanced RAG module with an adaptive query planning strategy and a triple-based structure ({\color{black}i.e.,} linking options, functions, and their dependencies) for the knowledge base. This module expands the LLM's accessible knowledge in an efficient and cost-effective manner. Also, it enables LLMs to better identify and interpret simulation functions, options, and their logical relationships than the standard RAG.
    \item {\color{black}It develops} an enhanced reasoning module by leveraging simulation-specific expertise, chain-of-thought prompting (CoT) and few-shot prompting. This module enables LLMs to fully understand their role, assigned tasks, reasoning pathways, and contextual knowledge (including retrieved information) in simulation tasks. Therefore, it strengthens their reasoning capabilities when generating simulation code.
    \item It further proposes a feedback-driven, multi-agent framework. It integrates the enhanced RAG and reasoning modules with an environmental interaction and error-correction mechanism. This framework facilitates both action execution and feedback reception. It provides responsive error signals to initiate adaptive adjustments for the RAG and reasoning modules to automatically correct errors, thereby enhancing the reliability of simulation outcomes. 
    \item Through testing across various strategies, simulation environments, and a diverse range of tasks, this paper reveals that even the latest LLM, o1-preview, struggles with power system simulation tasks, including those involving well-established tools like \LaTeXstyle{MATPOWER}, despite prior exposure in the pre-training of the LLM. This paper further reveals that high simulation success rates depend on the cumulative effect of multiple strategies. Following this idea, the proposed framework demonstrates high success rates, enabling cost-effective, rapid task completion. It, therefore, provides a scalable tool for power system researchers.
\end{itemize}

\FloatBarrier
\begin{figure*}[t!]
\centering
\includegraphics[width=1\linewidth]{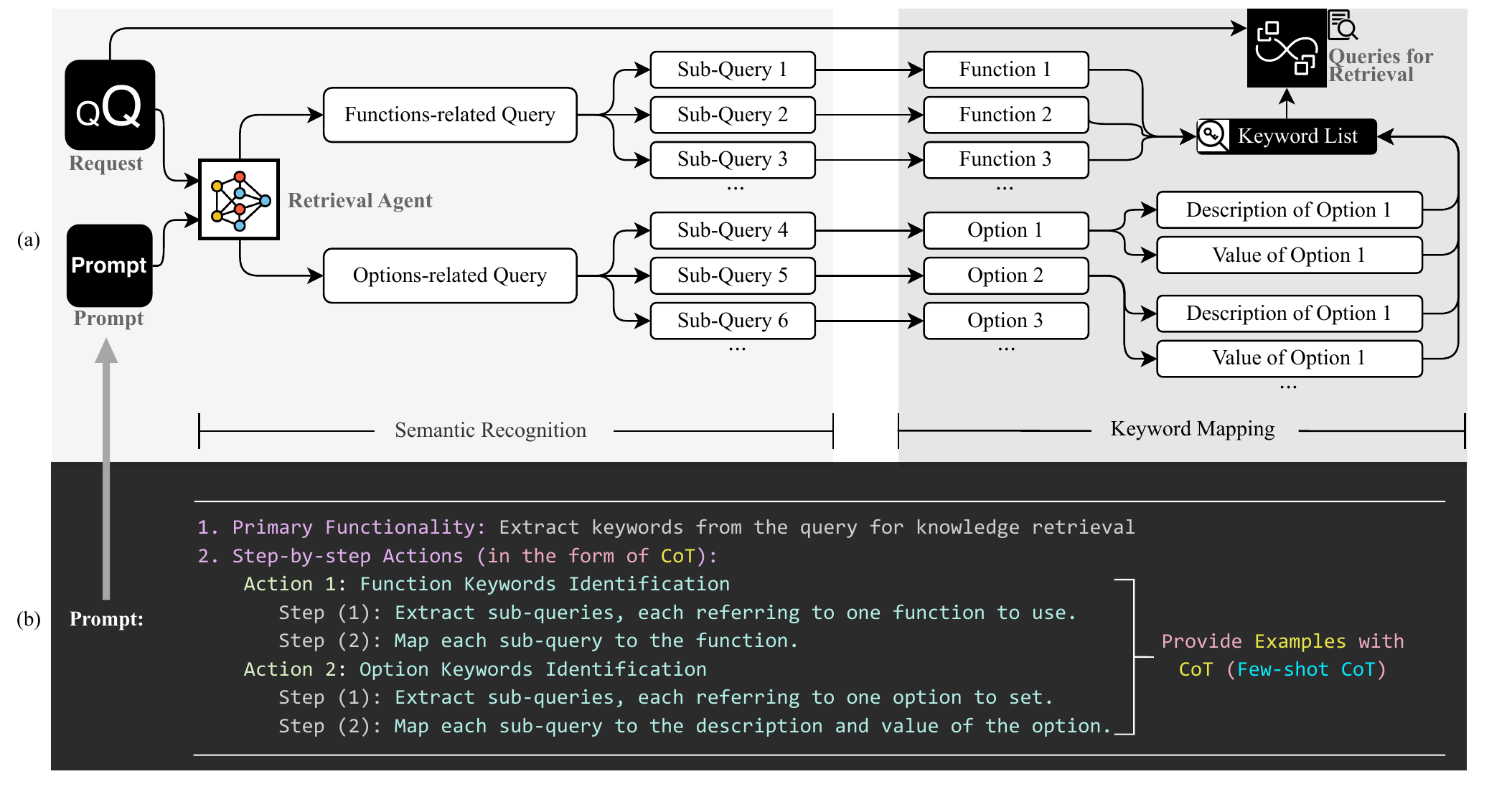} 
\caption{Enhanced RAG module for simulation tasks. (a) The retrieval agent decomposes simulation requests into function-related and option-related sub-queries, mapped to specific functions and options for precise keyword-based retrieval. (b) Structured prompt design detailing keyword extraction steps via few-shot CoT. }
\label{fig:enRAG}
\end{figure*}

\FloatBarrier
\begin{figure}[t!]
\centering
\includegraphics[width=1\linewidth]{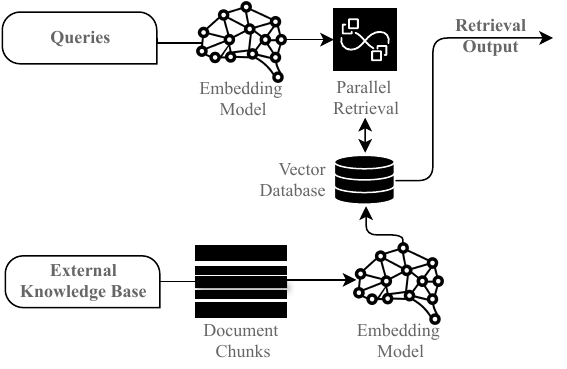} 
\caption{General RAG diagram, including the process of external knowledge chunking, text embedding, and parallel retrieval within a vector database to produce relevant retrieval output based on the input queries and the external knowledge base. Algorithm \ref{alg:embed} details the configuration settings and process for the embedding used in this paper.}
\label{fig:RAG}
\end{figure}

\begin{algorithm}[h]
    {\color{black}
    \caption{Embedding for External Documents}
    \label{alg:embed}
    \KwIn{\\
    \textit{datasetPath}: Paths to external documents \\
    \textit{chunkSize}: Words per chunk \\
    \textit{embeddingModel}: Chosen text embedding model \\
    \textit{apiKey}: Credential for embedding \\
    }
    \KwOut{ \\
        \textit{vectorDB}: Vector database for documents
    }
    \BlankLine
    
    \nonl \textcolor{gray}{// Step 1: Chunk}\;
    \textit{chunks} $\leftarrow$ \func{splitIntoChunks}{\textit{datasetPath}, \textit{chunkSize}}
    \BlankLine
    
    \nonl \textcolor{gray}{// Step 2: Embedding}\;
    \textit{vectors} $\leftarrow$ \func{embedChunks}{\textit{chunks}, \textit{embeddingModel}, \textit{apiKey}}
    \BlankLine
    
    \nonl \textcolor{gray}{// Step 3: Store in Vector DB}\;
    \textit{vectorDB} $\leftarrow$ \func{storeVectors}{\textit{vectors}, \textit{apiKey}}
    \BlankLine
    
    }
    \end{algorithm}

This paper, as a substantial extension of the {\color{black}authors'} preliminary work in \cite{jia2024enabling}, is structured as follows: Section II introduces the enhanced RAG module. Section III presents the enhanced reasoning module, and Section IV describes the environmental acting module with the feedback mechanism. Finally, Section V presents case study results, while Section VI concludes the paper with key findings and future outlook.

\section{Enhanced RAG Module}

As an efficient and scalable approach for integrating external knowledge to LLMs, RAG consists of three key steps: external knowledge chunking (splitting documents into smaller pieces), text embedding (converting texts into vectors using neural networks such as \texttt{text2vec}), and information retrieval (finding information in the vector space that aligns with the query) \cite{dong2024exploring}. Fig.~\ref{fig:RAG} illustrates a general RAG diagram. However, for power system simulations, critical questions arise: (i) \textit{what types of queries should be used for retrieval}? and (ii) \textit{what knowledge base should serve as the retrieval repository}? Addressing these questions reveals two primary areas for enhancing RAG's effectiveness in simulation tasks.

To this end, this paper proposes an enhanced RAG module. This is specifically designed to integrate power system simulation knowledge into LLMs and reduce hallucinations. This module emphasizes the identification of essential keywords in simulation requests to facilitate more precise knowledge retrieval than the standard RAG. It includes two main components: (i) an adaptive query planning strategy, and (ii) a triple-based structure design for the knowledge base. Together, these components provide an enhanced RAG for complex power system simulation tasks.


\subsection{Adaptive Query Planning}

This section addresses the question of what types of queries should be used for retrieval. In the standard RAG approach, the entire simulation request is processed as a single unit. As discussed and demonstrated in case studies, this conflates distinct elements in the request. As a result, it leads to inefficiencies and reduced retrieval accuracy. In fact, simulation requests typically contain two critical elements: the functions to be used and the options to be set. Thus, this paper proposes using functions and options as distinct retrieval queries. However, these elements are rarely stated explicitly in simulation requests; instead, they are embedded in natural language descriptions.

To address this, an agent-driven adaptive query planning strategy is developed, as shown in Fig.~\ref{fig:enRAG}(a), which automatically extracts function-related and option-related queries from the broader request to serve as retrieval keywords. The strategy operates in two phases: semantic recognition and keyword mapping, carried out by a retrieval agent (e.g., a general LLM). In the semantic recognition phase, the agent categorizes the request into two query types: function-related and option-related. Each function-related query is then decomposed into sub-queries, each corresponding to a potential simulation function to be used. Similarly, option-related queries are broken down into sub-queries, each linked to a potential option to be configured. This systematic separation ensures independent processing of each component within the request.

Following semantic recognition, the keyword mapping phase aligns each identified function and option sub-query with its precise keywords. For functions, the keywords are the functions' names. For options, sub-queries are further associated with their respective descriptions and values. Ultimately, the extracted functions, option descriptions, and values are entered as parallel retrieval queries.

To enable the retrieval agent to perform both semantic recognition and keyword mapping effectively, {\color{black}this paper designs} a structured, general action prompt that integrates chain-of-thought prompting (CoT) \cite{wei2022chain} and few-shot prompting \cite{mann2020language} (i.e., few-shot CoT), as depicted in Fig.~\ref{fig:enRAG}(b). Only the few-shot examples are tool-dependent, making them modular and easily adaptable. The rest of the prompt remains general and independent of specific simulation tools. 

{\color{black}
\noindent {\color{black}\textbf{Remark 1}}: The adaptive query planning strategy goes well beyond simple keyword extraction. It employs semantic recognition via few-shot CoT, which can translate descriptions into relevant keywords. This enables the retrieval agent to infer implicit simulation functions and options from context, even when explicit terms are missing. For example, consider the request: “\textit{For the IEEE 24-bus Reliability Test System, solve the AC optimal power flow (OPF) problem with the de-commitment of expensive generators},” which is the initial request in complex task 7 for \LaTeXstyle{MATPOWER} (see Section V for more details). Rather than extracting generic phrases like “AC optimal power flow” or “de-commitment,” the retrieval agent—guided by few-shot CoT prompting for function mapping—directly identifies “runuopf” as the keyword, i.e., the specific \LaTeXstyle{MATPOWER} function for OPF with de-commitment. 
}

\subsection{Triple-based Structure Design for Knowledge Base}

This section addresses the question of which knowledge base should serve as the retrieval repository. While each power system simulation tool includes a user manual with detailed instructions, this manual is not an ideal retrieval repository. The reasons are twofold: (i) User manuals are designed for human readability rather than automated retrieval; although readable, they are unstructured and inefficient for machine-driven queries, especially when manuals primarily consist of formulas, tables, and figures. (ii) The main challenge for LLMs in generating simulation code is understanding the logical dependencies between options and functions, as many options are function-dependent. Using only the user manual for retrieval fails to capture these complex relationships effectively.

To overcome these issues, {\color{black}this paper proposes} an additional, easy-to-construct retrieval repository: a triple-based structured option document. The following details the approach {\color{black}of building} such a document for a given simulation tool:
\begin{itemize}[leftmargin=10pt]
	\item \textbf{Step 1:} ChatGPT-4o is employed to parse the simulation manual and automatically extract key option entities. This process produces a list of options directly from the manual, including option names, potential default values/formats, and descriptions.
	\item \textbf{Step 2:} The same LLM is further utilized to analyze the textual context in which the options appear, thereby determining the logical relationships between each extracted option and its associated simulation function.
	\item \textbf{Step 3:} The outputs from Steps 1 and 2 are organized into a structured text document using a triple format, where each triple comprises: (i) the option name with its default value/format, (ii) the corresponding function dependency (linking each option and its related functions), and (iii) the description of the option along with the choices of its value. 
	\item \textbf{Step 4:} Finally, domain experts review and refine the automatically generated document to ensure that the logical associations between options and functions are accurate.
\end{itemize}
Note that the inclusion of the function dependency in the document enables retrieval for logical relationships. As will {\color{black}be demonstrated} in case studies, this supplementary repository significantly enhances retrieval efficiency and improves the accuracy of simulation code generation by preserving the logical context.

\subsection{Adaptability}

The enhanced RAG module demonstrates high adaptability to a wide range of power system simulation tools. This flexibility is achieved by capturing the core principles of simulation coding—the use of functions and options. Specifically, the module decomposes simulation requests into distinct queries related to functions and options. This decomposition is tool-independent, thereby enabling the seamless integration of new simulation platforms. Additionally, the construction of a triple-based option document—linking options to their corresponding functions and dependencies—leverages inherent logical relationships that are common across different simulation tools. Together, these design choices underscore the module's robust adaptability in diverse simulation environments.

\section{Enhanced Reasoning Module}

The enhanced RAG module provides LLMs with retrieval results tailored to a simulation request. However, it is still essential to strengthen the LLM's reasoning abilities. This ensures the generation of correct simulation codes based on the retrieval results. Hence, a coding agent is required (i.e., another LLM) to write {\color{black}code for simulation} tasks. This agent needs to fully understand its role, assigned tasks, reasoning path, and contextual knowledge, including retrieval results, when handling simulation tasks.

\begin{figure*}[t!]
    \centering
    \includegraphics[width=1\linewidth]{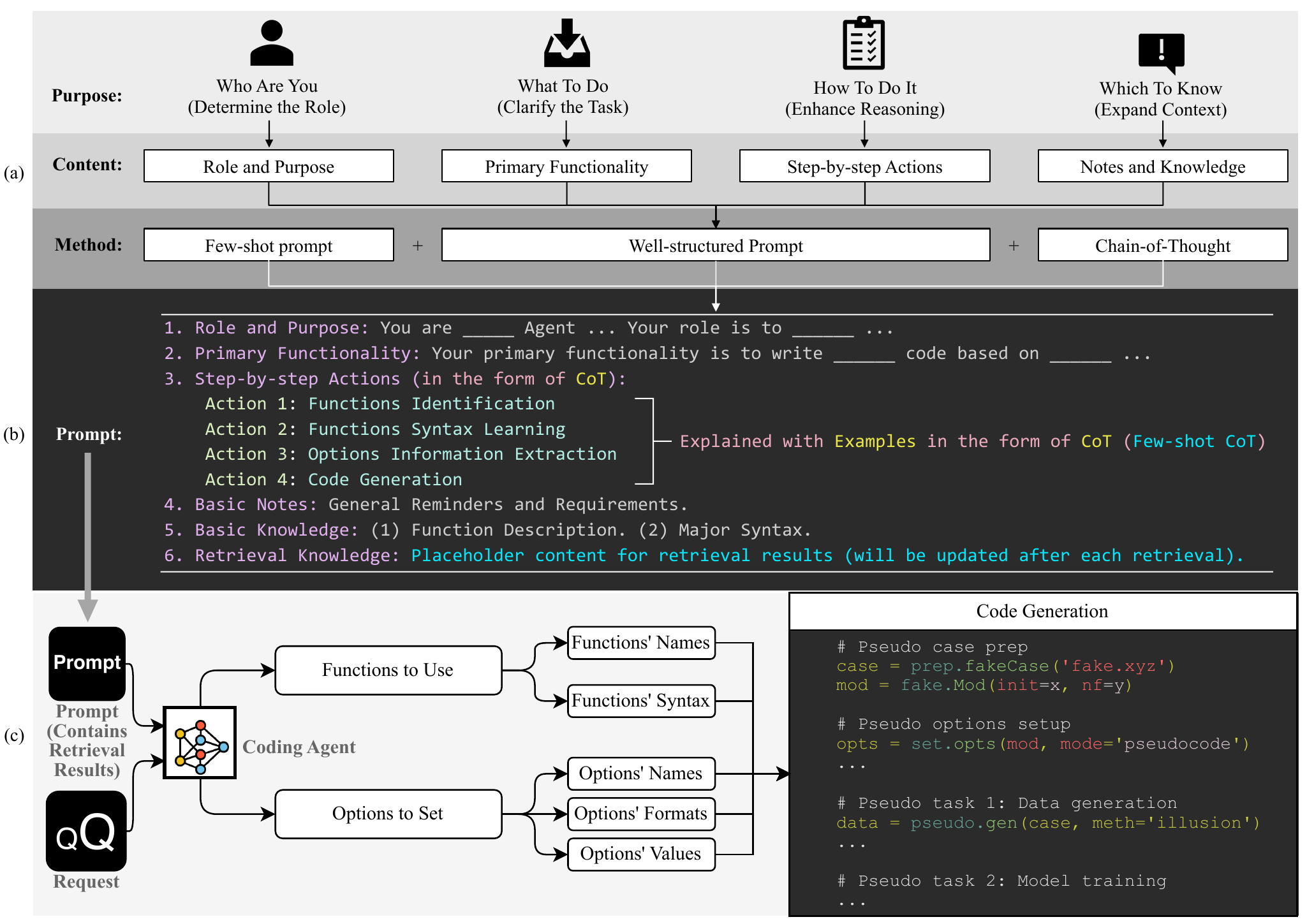} 
    \caption{Enhanced reasoning module for simulation code generation: (a) Core Concept, defining the coding agent’s role, assigned tasks, reasoning path, and contextual knowledge. (b) Structured prompt design equipped with few-shot CoT, in order to enhance the agent's reasoning ability for simulation. (c) Coding agent workflow integrating the designed prompt and simulation request to produce  simulation code.}
    \label{fig:reason}
\end{figure*}

To address this, {\color{black}this paper proposes} an enhanced reasoning module, as detailed in Fig.~\ref{fig:reason}. It provides structured guidance, sequential reasoning steps, and contextual knowledge to support accurate code generation by the coding agent. Details are as follows.  

\FloatBarrier 

\subsection{Role and Functionality Definition}

The agent deployed in this module is designated as a simulation coding agent for a specific simulation tool. Its primary function is to generate syntax-compliant simulation code that aligns with the specific task requirements, the static provided knowledge, and the dynamically retrieved knowledge.

\subsection{Reasoning Framework}

To enable systematic, tool-independent reasoning, {\color{black}this paper develops} a few-shot CoT framework, which breaks down the simulation task into the following universal actions:
\begin{itemize}[leftmargin=10pt] 
\item \textit{Function Identification}: Determines the functions relevant to the simulation task. \item \textit{Function Syntax Learning}: Acquires the correct syntax for identified functions to ensure compliance with the simulation tool’s requirements. 
\item \textit{Option Information Extraction}: Identifies options and extracts their formats, values, and dependencies to maintain coherence with the selected functions. 
\item \textit{Code Generation}: Integrates all extracted information into cohesive simulation code that meets task specifications and adheres to syntax requirements. 
\end{itemize}
Each of these actions is further clarified with tool-specific coding examples in the prompt. While the examples are tool-dependent, the rest of the framework remains general. Overall, the above reasoning framework highlights again that the key to handling simulation {\color{black}tasks is:} correctly identifying and combining functions and options.

\subsection{Knowledge Integration}

{\color{black}

The above reasoning actions heavily rely on information drawn from both the simulation request and supplementary knowledge, comprising:
\begin{itemize}[leftmargin=10pt]
    \item \textit{Static Basic Knowledge}: Supplies the agent with foundational information on essential functions and syntax rules pertinent to the simulation tool. This static knowledge serves as a base reference and reminder for the agent to consult when generating code. Note that such knowledge is tool-dependent. 
    \item \textit{Dynamic Retrieval Knowledge}: Complements static knowledge by incorporating real-time, request-specific details. This is achieved through the integration of the aforementioned RAG module, which retrieves relevant information on option formats, values, and function dependencies for accurate code generation. 
\end{itemize}

The integration of the RAG module into the LLM-based coding agent (from the reasoning module) is established through placeholders in the prompt. These placeholders serve as symbolic flags that are dynamically replaced by retrieval results before the prompt is sent to the coding agent. As shown in the ``Retrieval Knowledge'' section of Fig.~\ref{fig:reason}(b), during runtime, the enhanced RAG module retrieves simulation-specific information and substitutes these placeholders with up-to-date details on simulation functions, options, and parameters. Consequently, the coding agent generates simulation code based on both this dynamically retrieved information and the static knowledge embedded in the prompt.
}

Eventually, by integrating both static and dynamic knowledge, as well as the {\color{black}above-structured} reasoning framework, the coding agent is expected to generate accurate code to address the simulation request.

{\color{black}

\subsection{Adaptability}

The enhanced reasoning module leverages few-shot chain-of-thought prompting techniques, which embody universal principles and remain tool-agnostic. Additionally, supported by the RAG module, the reasoning module integrates two levels of knowledge: {\color{black}static} and dynamic knowledge. The static basic knowledge provides fundamental, coarse-grained insights, such as the basic principles of using a simulation tool. Meanwhile, {\color{black}the dynamically-retrieved} knowledge supplies detailed, fine-grained information. This dual-layer approach is universally applicable across different tools. As a result, the enhanced reasoning module adapts seamlessly to a variety of simulation tasks and environments.

}

\section{Environmental Acting Module with Feedback}

\begin{figure*}[t]
\centering
\includegraphics[width=1\linewidth]{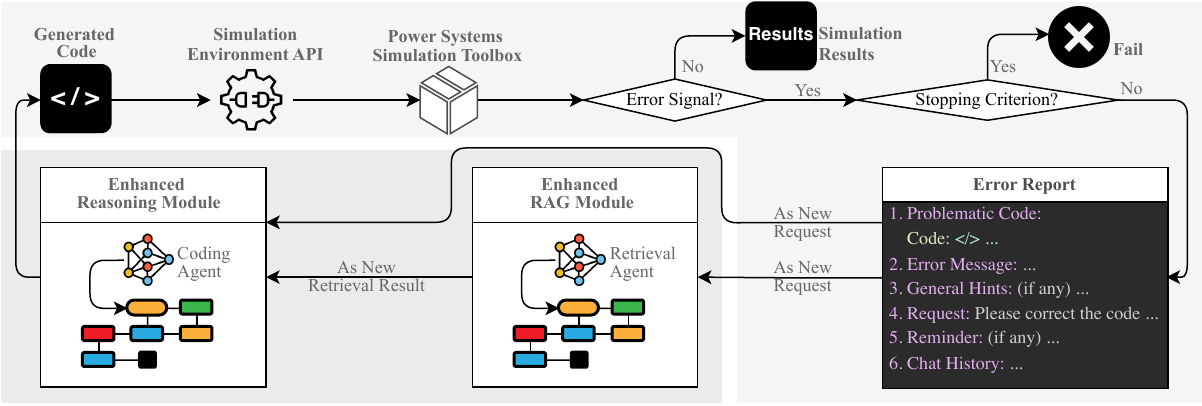} 
\caption{Environmental acting module with an {\color{black}error-feedback} mechanism that integrates with both the RAG and reasoning modules.}
\label{fig:act}
\end{figure*}

Despite the reinforcement brought by the enhanced RAG and reasoning modules, the coding agent may still encounter errors during simulation code generation. To address this, it is essential to enable direct interaction between the LLM and the simulation environment. It allows the agent to receive execution feedback and iteratively refine its code. To this end, {\color{black}this paper proposes} an environmental acting module with an {\color{black}error-feedback} mechanism that integrates with both the RAG and reasoning modules, as illustrated in Fig.~\ref{fig:act}. The components of this module are described {\color{black}below}. 

\subsection{Code Execution and Detection}

The simulation code, generated by the coding agent from the enhanced reasoning module, is executed using the simulation environment API connected to a 
 specific {\color{black}power system simulations} tool. Following execution, the simulation environment produces results, which are then checked for error signals. Specifically:
\begin{itemize}[leftmargin=10pt]
    \item If an error is detected, the code advances to a stopping criterion check. If the stopping criterion is met, the process is terminated; if not, the module triggers a feedback loop with detailed error reporting.  
     \item If no errors are detected, the process completes.
\end{itemize}

\subsection{Error Handling and Feedback Loop}

Upon detecting an error in the simulation results, an error report is automatically generated, containing:
\begin{itemize}[leftmargin=10pt]
    \item \textit{Problematic Code}: The code segment that caused the error.
    \item \textit{Error Message}: A detailed description of the error.
    \item \textit{General Hints}: Additional guidance on common issues.
    \item \textit{Request}: Specific corrections needed to address the error.
    \item \textit{Reminders}: Additional constraints or requirements, if any.
    \item \textit{Chat History}: A log of previous interactions and iterations.
\end{itemize}

    \begin{table*}[t]
    {\color{black}
    \captionsetup{justification=centerlast} 
    \caption{{\color{black}Evaluated schemes by distinct combinations of the proposed strategies; a checkmark indicates inclusion of the corresponding strategy. \\(GPT4o: APIs of gpt-4o-2024-05-13 and gpt-4o-2024-08-06, with the exact version specified in each evaluation; CGPT4o: ChatGPT4o; o1p: o1-preview)}}
    \centering
    \footnotesize 
    \label{tab:strategy}
    \renewcommand{\arraystretch}{1} 
    \setlength{\tabcolsep}{1.5pt}
    \begin{tabular}{lllllllllllllll}
    \toprule 
                                                           & \multicolumn{1}{c}{\shortstack{GPT4o \\ Full}}                & \multicolumn{1}{c}{\shortstack{GPT4o \\ PR}}                  & \multicolumn{1}{c}{\shortstack{GPT4o \\ RSR}}                 & \multicolumn{1}{c}{\shortstack{GPT4o \\ SR}}                  & \multicolumn{1}{c}{\shortstack{GPT4o \\ Sole}}                & \multicolumn{1}{c}{\shortstack{GPT4o \\ NC}}                  & \multicolumn{1}{c}{\shortstack{GPT4o \\ NP}}                  & \multicolumn{1}{c}{\shortstack{GPT4o \\ NS}}                  & \multicolumn{1}{c}{\shortstack{GPT4o \\ NR}}                  & \multicolumn{1}{c}{\shortstack{GPT4o \\ NCS}}                 & \multicolumn{1}{c}{\shortstack{GPT4o \\ RSRNW}}               & \multicolumn{1}{c}{\shortstack{CGPT4o \\ R}}                  & \multicolumn{1}{c}{\shortstack[c]{o1p\\ Sole}}     & \multicolumn{1}{c}{\shortstack{GPT4o \\ Sole-SFT}}               \\
    \cellcolor[HTML]{D6D6D6}Query Planning  & \multicolumn{1}{c}{\cellcolor[HTML]{D6D6D6}\checkmark} & \multicolumn{1}{c}{\cellcolor[HTML]{D6D6D6}\checkmark} & \multicolumn{1}{c}{\cellcolor[HTML]{D6D6D6}}  & \multicolumn{1}{c}{\cellcolor[HTML]{D6D6D6}}  & \multicolumn{1}{c}{\cellcolor[HTML]{D6D6D6}}  & \multicolumn{1}{c}{\cellcolor[HTML]{D6D6D6}\checkmark} & \multicolumn{1}{c}{\cellcolor[HTML]{D6D6D6}\checkmark} & \multicolumn{1}{c}{\cellcolor[HTML]{D6D6D6}\checkmark} & \multicolumn{1}{c}{\cellcolor[HTML]{D6D6D6}}  & \multicolumn{1}{c}{\cellcolor[HTML]{D6D6D6}\checkmark} & \multicolumn{1}{c}{\cellcolor[HTML]{D6D6D6}\checkmark} & \multicolumn{1}{c}{\cellcolor[HTML]{D6D6D6}}  & \multicolumn{1}{c}{\cellcolor[HTML]{D6D6D6}} & \multicolumn{1}{c}{\cellcolor[HTML]{D6D6D6}} \\
    \cellcolor[HTML]{D9D9D9}Triple-based Structured Option Document & \multicolumn{1}{c}{\cellcolor[HTML]{D6D6D6}\checkmark} & \multicolumn{1}{c}{\cellcolor[HTML]{D6D6D6}\checkmark} & \multicolumn{1}{c}{\cellcolor[HTML]{D6D6D6}\checkmark} & \multicolumn{1}{c}{\cellcolor[HTML]{D6D6D6}\checkmark} & \multicolumn{1}{c}{\cellcolor[HTML]{D6D6D6}}  & \multicolumn{1}{c}{\cellcolor[HTML]{D6D6D6}\checkmark} & \multicolumn{1}{c}{\cellcolor[HTML]{D6D6D6}}  & \multicolumn{1}{c}{\cellcolor[HTML]{D6D6D6}\checkmark} & \multicolumn{1}{c}{\cellcolor[HTML]{D6D6D6}}  & \multicolumn{1}{c}{\cellcolor[HTML]{D6D6D6}\checkmark} & \multicolumn{1}{c}{\cellcolor[HTML]{D6D6D6}\checkmark} & \multicolumn{1}{c}{\cellcolor[HTML]{D6D6D6}\checkmark} & \multicolumn{1}{c}{\cellcolor[HTML]{D6D6D6}} & \multicolumn{1}{c}{\cellcolor[HTML]{D6D6D6}} \\
    
    \cellcolor[HTML]{D6D6D6}Chain of Thought   Prompting             & \multicolumn{1}{c}{\cellcolor[HTML]{D6D6D6}\checkmark} & \multicolumn{1}{c}{\cellcolor[HTML]{D6D6D6}}  & \multicolumn{1}{c}{\cellcolor[HTML]{D6D6D6}\checkmark} & \multicolumn{1}{c}{\cellcolor[HTML]{D6D6D6}}  & \multicolumn{1}{c}{\cellcolor[HTML]{D6D6D6}}  & \multicolumn{1}{c}{\cellcolor[HTML]{D6D6D6}}  & \multicolumn{1}{c}{\cellcolor[HTML]{D6D6D6}\checkmark} & \multicolumn{1}{c}{\cellcolor[HTML]{D6D6D6}\checkmark} & \multicolumn{1}{c}{\cellcolor[HTML]{D6D6D6}\checkmark} & \multicolumn{1}{c}{\cellcolor[HTML]{D6D6D6}}  & \multicolumn{1}{c}{\cellcolor[HTML]{D6D6D6}\checkmark} & \multicolumn{1}{c}{\cellcolor[HTML]{D6D6D6}}  & \multicolumn{1}{c}{\cellcolor[HTML]{D6D6D6}} & \multicolumn{1}{c}{\cellcolor[HTML]{D6D6D6}} \\
    \cellcolor[HTML]{D6D6D6}Few-Shot Prompting                       & \multicolumn{1}{c}{\cellcolor[HTML]{D6D6D6}\checkmark} & \multicolumn{1}{c}{\cellcolor[HTML]{D6D6D6}}  & \multicolumn{1}{c}{\cellcolor[HTML]{D6D6D6}\checkmark} & \multicolumn{1}{c}{\cellcolor[HTML]{D6D6D6}}  & \multicolumn{1}{c}{\cellcolor[HTML]{D6D6D6}}  & \multicolumn{1}{c}{\cellcolor[HTML]{D6D6D6}\checkmark} & \multicolumn{1}{c}{\cellcolor[HTML]{D6D6D6}\checkmark} & \multicolumn{1}{c}{\cellcolor[HTML]{D6D6D6}}  & \multicolumn{1}{c}{\cellcolor[HTML]{D6D6D6}\checkmark} & \multicolumn{1}{c}{\cellcolor[HTML]{D6D6D6}}  & \multicolumn{1}{c}{\cellcolor[HTML]{D6D6D6}\checkmark} & \multicolumn{1}{c}{\cellcolor[HTML]{D6D6D6}}  & \multicolumn{1}{c}{\cellcolor[HTML]{D6D6D6}}  & \multicolumn{1}{c}{\cellcolor[HTML]{D6D6D6}} \\
    \cellcolor[HTML]{D6D6D6}Static Basic Knowledge                   & \multicolumn{1}{c}{\cellcolor[HTML]{D6D6D6}\checkmark} & \multicolumn{1}{c}{\cellcolor[HTML]{D6D6D6}}  & \multicolumn{1}{c}{\cellcolor[HTML]{D6D6D6}\checkmark} & \multicolumn{1}{c}{\cellcolor[HTML]{D6D6D6}}  & \multicolumn{1}{c}{\cellcolor[HTML]{D6D6D6}}  & \multicolumn{1}{c}{\cellcolor[HTML]{D6D6D6}\checkmark} & \multicolumn{1}{c}{\cellcolor[HTML]{D6D6D6}\checkmark} & \multicolumn{1}{c}{\cellcolor[HTML]{D6D6D6}\checkmark} & \multicolumn{1}{c}{\cellcolor[HTML]{D6D6D6}\checkmark} & \multicolumn{1}{c}{\cellcolor[HTML]{D6D6D6}\checkmark} & \multicolumn{1}{c}{\cellcolor[HTML]{D6D6D6}\checkmark} & \multicolumn{1}{c}{\cellcolor[HTML]{D6D6D6}}  & \multicolumn{1}{c}{\cellcolor[HTML]{D6D6D6}}  & \multicolumn{1}{c}{\cellcolor[HTML]{D6D6D6}} \\
    
    \cellcolor[HTML]{D6D6D6}Environmental Acting and Feedback      & \multicolumn{1}{c}{\cellcolor[HTML]{D6D6D6}\checkmark} & \multicolumn{1}{c}{\cellcolor[HTML]{D6D6D6}\checkmark} & \multicolumn{1}{c}{\cellcolor[HTML]{D6D6D6}\checkmark} & \multicolumn{1}{c}{\cellcolor[HTML]{D6D6D6}\checkmark} & \multicolumn{1}{c}{\cellcolor[HTML]{D6D6D6}\checkmark} & \multicolumn{1}{c}{\cellcolor[HTML]{D6D6D6}\checkmark} & \multicolumn{1}{c}{\cellcolor[HTML]{D6D6D6}\checkmark} & \multicolumn{1}{c}{\cellcolor[HTML]{D6D6D6}\checkmark} & \multicolumn{1}{c}{\cellcolor[HTML]{D6D6D6}\checkmark} & \multicolumn{1}{c}{\cellcolor[HTML]{D6D6D6}\checkmark} & \multicolumn{1}{c}{\cellcolor[HTML]{D6D6D6}\checkmark} & \multicolumn{1}{c}{\cellcolor[HTML]{D6D6D6}\checkmark} & \multicolumn{1}{c}{\cellcolor[HTML]{D6D6D6}\checkmark} & \multicolumn{1}{c}{\cellcolor[HTML]{D6D6D6}\checkmark} \\
    
    \cellcolor[HTML]{D6D6D6}Proposed RAG                             & \multicolumn{1}{c}{\cellcolor[HTML]{D6D6D6}\checkmark} & \multicolumn{1}{c}{\cellcolor[HTML]{D6D6D6}\checkmark} & \multicolumn{1}{c}{\cellcolor[HTML]{D6D6D6}}  & \multicolumn{1}{c}{\cellcolor[HTML]{D6D6D6}}  & \multicolumn{1}{c}{\cellcolor[HTML]{D6D6D6}}  & \multicolumn{1}{c}{\cellcolor[HTML]{D6D6D6}\checkmark} & \multicolumn{1}{c}{\cellcolor[HTML]{D6D6D6}\checkmark} & \multicolumn{1}{c}{\cellcolor[HTML]{D6D6D6}\checkmark} & \multicolumn{1}{c}{\cellcolor[HTML]{D6D6D6}}  & \multicolumn{1}{c}{\cellcolor[HTML]{D6D6D6}\checkmark} & \multicolumn{1}{c}{\cellcolor[HTML]{D6D6D6}\checkmark} & \multicolumn{1}{c}{\cellcolor[HTML]{D6D6D6}}  & \multicolumn{1}{c}{\cellcolor[HTML]{D6D6D6}} & \multicolumn{1}{c}{\cellcolor[HTML]{D6D6D6}} \\
    \cellcolor[HTML]{FFFFFF}Standard RAG                             & \multicolumn{1}{c}{}                          & \multicolumn{1}{c}{}                          & \multicolumn{1}{c}{\checkmark}                         & \multicolumn{1}{c}{\checkmark}                         & \multicolumn{1}{c}{}                          & \multicolumn{1}{c}{}                          & \multicolumn{1}{c}{}                          & \multicolumn{1}{c}{}                          & \multicolumn{1}{c}{}                          & \multicolumn{1}{c}{}                          & \multicolumn{1}{c}{}                          & \multicolumn{1}{c}{}                          & \multicolumn{1}{c}{}                        \\
    \cellcolor[HTML]{FFFFFF}OpenAI's Built-in RAG                    & \multicolumn{1}{c}{}                          & \multicolumn{1}{c}{}                          & \multicolumn{1}{c}{}                          & \multicolumn{1}{c}{}                          & \multicolumn{1}{c}{}                          & \multicolumn{1}{c}{}                          & \multicolumn{1}{c}{}                          & \multicolumn{1}{c}{}                          & \multicolumn{1}{c}{}                          & \multicolumn{1}{c}{}                          & \multicolumn{1}{c}{}                          & \multicolumn{1}{c}{\checkmark}                         & \multicolumn{1}{c}{}            & \multicolumn{1}{c}{}             \\
    \cellcolor[HTML]{FFFFFF}OpenAI's Built-in Supervised Fine-tuning                    & \multicolumn{1}{c}{}                          & \multicolumn{1}{c}{}                          & \multicolumn{1}{c}{}                          & \multicolumn{1}{c}{}                          & \multicolumn{1}{c}{}                          & \multicolumn{1}{c}{}                          & \multicolumn{1}{c}{}                          & \multicolumn{1}{c}{}                          & \multicolumn{1}{c}{}                          & \multicolumn{1}{c}{}                          & \multicolumn{1}{c}{}                          & \multicolumn{1}{c}{}                         & \multicolumn{1}{c}{}            & \multicolumn{1}{c}{\checkmark}             \\
    
                                                 Well-developed   Error-reporting System                          & \multicolumn{1}{c}{\checkmark}                         & \multicolumn{1}{c}{\checkmark}                         & \multicolumn{1}{c}{\checkmark}                         & \multicolumn{1}{c}{\checkmark}                         & \multicolumn{1}{c}{\checkmark}                         & \multicolumn{1}{c}{\checkmark}                         & \multicolumn{1}{c}{\checkmark}                         & \multicolumn{1}{c}{\checkmark}                         & \multicolumn{1}{c}{\checkmark}                         & \multicolumn{1}{c}{\checkmark}                         & \multicolumn{1}{c}{}                          & \multicolumn{1}{c}{\checkmark}                         & \multicolumn{1}{c}{\checkmark}                      & \multicolumn{1}{c}{\checkmark}   \\
    \bottomrule 
    \end{tabular}
    }
    \end{table*}

\subsection{Enhanced RAG and Reasoning Module Interplay}

The error report and feedback are then processed as a new request by the retrieval agent in the enhanced RAG module. This agent retrieves relevant information based on the error report (the query planning can also be applied to error reporting by simply replacing the identification of function/option keywords with the identification of error-related keywords). The retrieved information is then passed to the enhanced reasoning module. The coding agent there uses both the retrieval results and the correction request to revise the simulation code. The updated code is fed back into the environmental acting module. This loop continues until the code meets all requirements or the stopping criterion is reached.

This design achieves two main objectives. First, the system continuously monitors simulation executions and triggers adaptive adjustments as needed, ensuring that all module interdependencies are effectively managed. Second, it handles failure modes. The error-feedback mechanism detects failures in both the RAG and reasoning modules—failures that can lead to simulation errors. Once a failure is detected, the mechanism initiates iterative corrective actions using adaptive prompts, which trigger updated retrievals and adjustments, thereby dynamically managing module failures. This conclusion is further reinforced by the case studies in Section V.

{\color{black}

\subsection{Adaptability}

The environmental acting module features a tool-independent design. First, it leverages the error detection and reporting systems common to many power system simulation tools. Second, it formulates the {\color{black}error-feedback} loop in a general manner, as detailed in Section IV-B. Each component is independent of any specific simulation tool. This design not only supports automatic error correction but also {\color{black}improves} adaptability to diverse simulation platforms.

}

\subsection{Pseudocode for the Full Framework}


Algorithm \ref{alg:full} presents the complete pseudocode for the proposed framework. The algorithm employs self-explanatory pseudo-functions and pseudo-options to illustrate the core modules, including their internal processes and the interplay among them. 

\noindent {\color{black}\textbf{Remark 2}}: It is important to emphasize that the proposed framework is built on the premise that simulation tools are learnable by humans through sufficient documentation. If a human can use a simulation tool based on its manuals, then the proposed integrated RAG, reasoning, and feedback modules can replicate and enhance this process to improve efficiency. Hence, for scenarios where documentation is absent, it is suggested to generate necessary documents, possibly with LLMs' help. These documents can then be incorporated into the proposed framework.

\section{Case Study}

To comprehensively validate the proposed framework, {\color{black}a range of tests have been carried out, differing in three key dimensions}: (i) distinct combinations of the proposed strategies within the framework to evaluate each strategy's independent effectiveness; (ii) different simulation environments, specifically \LaTeXstyle{Daline} \cite{Daline} and \LaTeXstyle{MATPOWER} \cite{zimmerman2010matpower}, which include tools both familiar and unfamiliar to LLMs\footnote{\LaTeXstyle{Daline} is available \href{https://www.shuo.science/daline}{[\underline{here}]} with a user manual in \cite{DalineM}. \LaTeXstyle{MATPOWER} is available \href{https://matpower.org/}{[\underline{here}]} with a user manual in \cite{zimmerman2024matpower}.}, to demonstrate the framework's versatility across various applications; and (iii) a wide array of simulation tasks, spanning normal to complex scenarios, to assess the framework's performance across various simulation demands.

The following sections detail the case study configurations, followed by an analysis of the simulation outcomes for \LaTeXstyle{Daline} and \LaTeXstyle{MATPOWER}. Eventually, the cost of using LLMs to perform power system simulations is discussed. All supporting materials, including prompts, knowledge bases, simulation tasks, training datasets, and results {\color{black}(i.e., the generated codes and their benchmarks, totaling 870 coding files) will be openly available upon acceptance.}


\begin{algorithm}[htbp]
    {\color{black}
    \caption{Feedback-driven Multi-agent Framework}
    \label{alg:full}
    \KwIn{\ 
    \textit{modelVer}: Model version from Table \ref{tab:strategy} \\
    \textit{config}: Configuration settings from Table \ref{tab:llm_rag_config} \\
    \textit{ragPrompt}: Query planning prompt from Fig.~\ref{fig:enRAG} \\
    \textit{reasonPrompt}: Structured reasoning prompt from Fig.~\ref{fig:reason} \\
    \textit{errTemplate}: Error report template from Fig.~\ref{fig:act} \\
    \textit{vectorDB}: Vector database from Algorithm \ref{alg:embed} \\
    \textit{task}: Simulation task \\
    \textit{maxAttempts}: Max attempt number \\
    \textit{apiKey}: Credentials for LLMs \\
    \textit{returnNum}: Number of top-relevant chunks to return
    }
    \KwOut{\
    \textit{simResult}: Simulation result \\
    \textit{code}: Generated code}
    \BlankLine
    
    \nonl \textcolor{gray}{// Preparation}\;
    \textit{attemptTime} $\leftarrow$ 0\;
    \textit{chatHistory} $\leftarrow$ emptySet\;
    \textit{env} $\leftarrow$ \func{activateSimulationEnvironment}{}
    \BlankLine
    
    \nonl \textcolor{gray}{// Enhanced RAG Module}\;
    \textit{retrievalLLM} $\leftarrow$ \func{setupLLM}{\textit{config}, \textit{modelVer}, \textit{ragPrompt}, \textit{apiKey}}\;
    \textit{keywords} $\leftarrow$ \func{generateKeywords}{\textit{retrievalLLM}, \textit{task}}\;
    \textit{retrievalInfo} $\leftarrow$ \func{parallelRetrieval}{\textit{retrievalLLM}, \textit{keywords}, \textit{vectorDB}, \textit{returnNum}}\;
    \BlankLine
    
    \nonl \textcolor{gray}{// Enhanced Reasoning Module}\;
    \textit{reasonPrompt} $\leftarrow$ \func{insertRetrieval}{\textit{reasonPrompt}, \textit{retrievalInfo}}\;
    \textit{codeLLM} $\leftarrow$ \func{setupLLM}{\textit{config}, \textit{modelVer}, \textit{reasonPrompt}, \textit{apiKey}}\;
    [\textit{code}, \textit{chatHistory}] $\leftarrow$ \func{generateCode}{\textit{codeLLM}, \textit{task}, \textit{chatHistory}}\;
    \BlankLine
    
    \nonl \textcolor{gray}{// Environmental Acting Module with Feedback}\;
    [\textit{simResult}, \textit{err}] $\leftarrow$ \func{runSimulation}{\textit{code}, \textit{env}}\;
    \BlankLine
    
    \While{\textit{err} $\neq$ null \textbf{and} \textit{attemptTime} $\leq$ \textit{maxAttempts}}{
        \textit{errReport} $\leftarrow$ \func{generateErrorReport}{\textit{code}, \textit{err}, \textit{errTemplate}}\;
        \textit{keywords} $\leftarrow$ \func{generateKeywords}{\textit{retrievalLLM}, \textit{errReport}}\;
        \textit{newRetrievalInfo} $\leftarrow$ \func{parallelRetrieval}{\textit{retrievalLLM}, \textit{keywords}, \textit{vectorDB}, \textit{returnNum}}\;
        \textit{reasonPrompt} $\leftarrow$ \func{insertRetrieval}{\textit{reasonPrompt}, \textit{newRetrievalInfo}}\;
        \textit{codeLLM} $\leftarrow$ \func{setupLLM}{\textit{config}, \textit{modelVer}, \textit{reasonPrompt}, \textit{apiKey}}\;
        [\textit{code}, \textit{chatHistory}] $\leftarrow$ \func{generateCode}{\textit{codeLLM}, \textit{errReport}, \textit{chatHistory}}\;
        [\textit{simResult}, \textit{err}] $\leftarrow$ \func{runSimulation}{\textit{code}, \textit{env}}\;
        \textit{attemptTime} $\leftarrow$ \textit{attemptTime} + 1\;
    }
    \BlankLine
    
    \If{\textit{err} $\neq$ null}{
        \func{reportFailure}{\textit{code}}\;
        \func{saveFailedCode}{\textit{code}}\;
    }
    \Else{
        \func{outputResult}{\textit{simResult}}\;
        \func{saveSimulationCode}{\textit{code}}\;
    }
    }
\end{algorithm}

\subsection{Settings}

Firstly, Table \ref{tab:strategy} presents the distinct combinations of the proposed strategies within the framework employed in the evaluation, with the proposed strategies shaded in gray. The model versions of LLMs can also be found in the caption of Table \ref{tab:strategy}, as well as in the caption of each evaluation figure/table. Meanwhile, the configurations used in all the evaluations are detailed in Table \ref{tab:llm_rag_config}.

\begin{table}[h]
    \centering
    {\color{black}
    \footnotesize
    \caption{{\color{black} Configuration settings for evaluations (for the LLM model versions, please see Table I or the caption of each evaluation figure/table)}}
    \vspace{-2pt}
    \renewcommand{\arraystretch}{1.2} 
    \setlength{\tabcolsep}{0.8pt}
    \label{tab:llm_rag_config}
    \begin{tabular}{cc>{\centering\arraybackslash}p{5.5cm}}
    \hline
    \multicolumn{3}{c}{\textbf{RAG Configuration}} \\ \hline
    \textbf{Parameter} & \textbf{Value} & \textbf{Description} \\ \hline
    Chunk size for .txt & 30 & Number of words per chunk for text files  \\ 
    Chunk size for .pdf & 50 & Number of words per chunk for PDF files \\ 
    Embedding Model & \texttt{v1} & See \href{https://help.aliyun.com/zh/dashscope/developer-reference/text-embedding-quick-start?spm=a2c4g.11186623.0.0.5695f97eD8MhdE}{[\underline{here}]} for the model \texttt{text-embedding-v1} \\ 
    Return number & 20 & Number of top-relevant chunks to return \\ 
    \hline \hline
    \multicolumn{3}{c}{\textbf{LLM Configuration}} \\ \hline
    \textbf{Parameter} & \textbf{Value} & \textbf{Description} \\ \hline
    Temperature & 0.1 & Randomness (lower = more predictable) \\ \hline
    Max Tokens & 4096 & Max. length of the output in tokens \\ \hline
    Top P & 1 & Probability threshold for token selection \\ \hline
    Frequency Penalty & 0 & Zero means common words are not suppressed \\ \hline
    Presence Penalty & 0 & Zero allows natural repetition when needed \\ \hline
    \hline \hline
    \multicolumn{3}{c}{\textbf{Acting Configuration}} \\ \hline
    \textbf{Parameter} & \textbf{Value} & \textbf{Description} \\ \hline
    \LaTeXstyle{Daline} $N_{\max}^t$  & 3 & Max. attempt number for \LaTeXstyle{Daline} task $t$ \\ \hline
    \LaTeXstyle{MATPOWER}  $N_{\max}^t$ & 5 & Max. attempt number for \LaTeXstyle{MATPOWER} task $t$ \\ \hline
    \end{tabular}}
\end{table}


Secondly, this paper selects \LaTeXstyle{Daline} \cite{Daline} and \LaTeXstyle{MATPOWER} \cite{zimmerman2010matpower} as the simulation environments. It is important to note that \LaTeXstyle{Daline} was released after the latest updates of the LLMs used in the evaluation, while the well-established tool \LaTeXstyle{MATPOWER} was already included in the training dataset of the LLMs. Consequently, these two environments encompass both seen and unseen scenarios for the LLMs, allowing us to demonstrate the framework's versatility.  

\begin{figure*}[t!]
    \centering
    \includegraphics[width=1\linewidth]{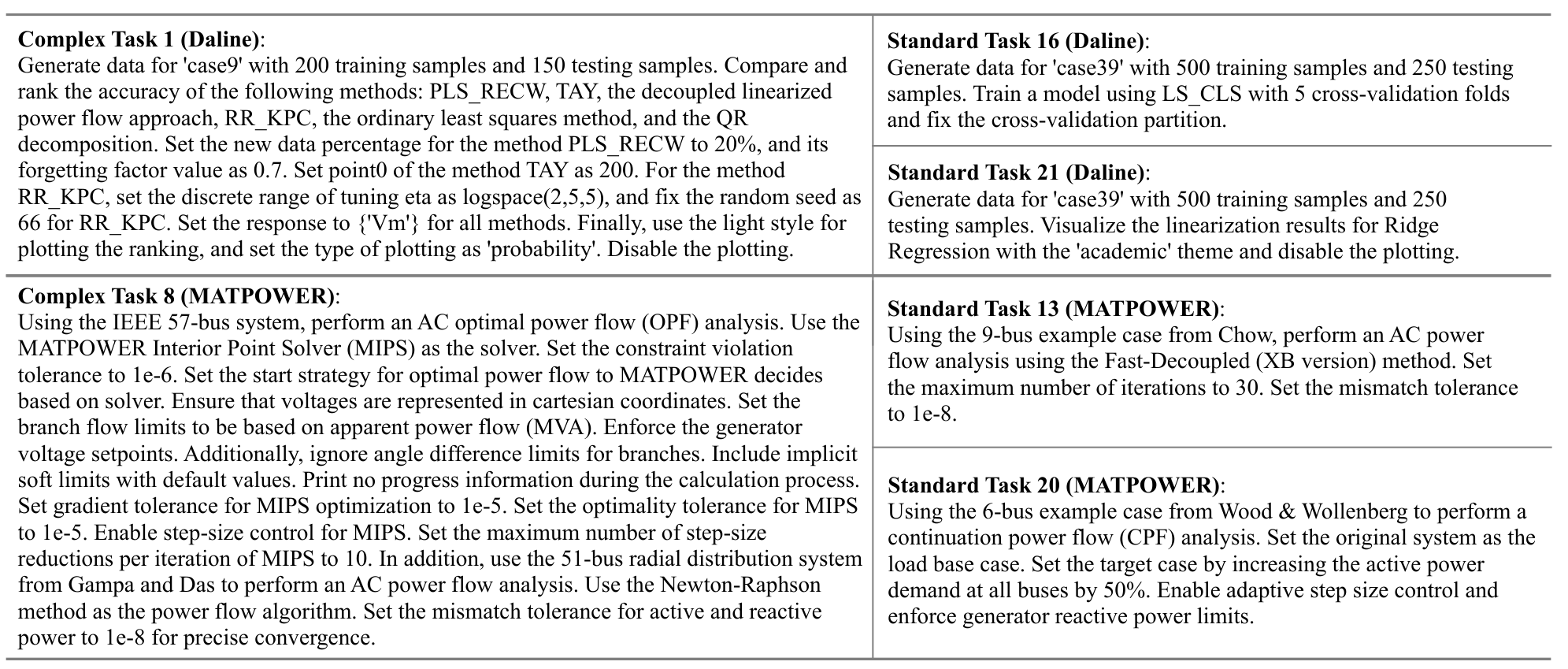} 
    \caption{Representative examples of simulation tasks for \LaTeXstyle{Daline} and  \LaTeXstyle{MATPOWER}.}
    \vspace{-6mm}
    \label{fig:task}
    \end{figure*}

{\color{black}
Thirdly, 34 simulation tasks were used to evaluate the proposed framework on  \LaTeXstyle{Daline}. Similarly, for \LaTeXstyle{MATPOWER}, 35 simulation tasks have been defined, including 8 complex tasks and 27 standard tasks. 
\begin{itemize}[leftmargin=10pt]
    \item The simulation tasks of \LaTeXstyle{Daline} span from basic data generation to advanced workflows that integrate data creation, data corruption, noise handling, outlier filtering, model training, method comparison, result visualization, and full-cycle management. The simulation suite exercises all distinct functions in \LaTeXstyle{Daline} across various power system cases. The tasks employ multiple modeling methods (e.g., Least Squares with Huber Weighting Function, Partial Least Squares with Clustering, Ridge Regression with Voltage-angle Coupling, Locally Weighted Ridge Regression, State-independent Voltage-angle Decoupled Method with Data Correction, to name a few) and a wide array of parameters (sample sizes, base types, noise levels, outlier techniques, cross-validation settings, method hyperparameters, etc.). 
    \item The simulation tasks of \LaTeXstyle{MATPOWER} span standard AC and DC optimal power flow (or just load flow) analyses as well as advanced continuation power flow (CPF) studies. They employ a variety of solvers and algorithms—including Newton-Raphson, Fast-Decoupled (both XB and BX versions), Gauss-Seidel, Implicit Z-bus Gauss, MIPS, fmincon, and GUROBI—across numerous test cases such as the IEEE 9-, 14-, 30-, 57-, 118-, and 145-bus systems, along with specialized cases like the 4-bus, 6-bus, 39-bus New England, 51-bus radial, 60-bus Nordic, and IEEE RTS 24-bus systems. Key parameters such as maximum iterations, mismatch tolerances, coordinate representations, and branch flow constraints were systematically varied. In the CPF analyses, both natural and pseudo arc length parameterizations were explored with adaptive step sizing, explicit nose point detection, and generation/load scaling. Additional configurations addressed generator reactive power limits, voltage setpoints, alternative formulations (current versus power balance), and solver-specific settings like gradient, optimality, and termination tolerances. 
\end{itemize}
Overall, these tasks comprehensively cover both basic and advanced functionalities of \LaTeXstyle{Daline} and \LaTeXstyle{MATPOWER} across a broad range of power system analysis scenarios. A selection of representative tasks is given in Fig.~\ref{fig:task}, intended as an exemplary overview.

}


Finally, to evaluate each scheme's performance, we compute a success rate based on the points obtained across multiple simulation tasks. Let \( T \) denote the total number of simulation tasks. For each task \( t \) (\(1 \leq t \leq T\)), let \( N_{\max}^{(t)} \) be the maximum number of allowed attempts for that task and let \( P_{t,i} \) be the points awarded on attempt \( i \) (with \( 1 \leq i \leq N_{\max}^{(t)} \)). The scoring for task $t$ at attempt $i$ is defined as follows:
{\color{black}
\begin{itemize}[leftmargin=10pt]
    \item \( P_{t,i} = 100 \) points if all requirements are fully satisfied and no irrelevant settings appear in the code.
    \item \( P_{t,i} = 50 \) points if all requirements are satisfied, but the code includes irrelevant or unnecessary settings that do not affect task completion.
    \item \( P_{t,i} = 0 \) points if any requirement is not satisfied.
\end{itemize}
}
Subsequent attempts are made only if the previous attempt resulted in an error, and any unused attempts are assigned the same score as the last executed attempt. Thus, the total score for task \( t \) is given by:
\[
S_t = \sum\nolimits_{i=1}^{N_{\max}^{(t)}} P_{t,i},
\]
with a maximum possible score of \( 100 \times N_{\max}^{(t)} \) for that task. 
Finally, the overall success rate across all tasks is computed as:
\[
R_{\text{overall}} = \frac{\sum_{t=1}^{T} S_t}{\sum_{t=1}^{T} \left(100 \times N_{\max}^{(t)}\right)} \times 100\%.
\]
In the following experiments, \( N_{\max}^t = 3 \) ($\forall t$) is set for \LaTeXstyle{Daline} tasks, and \( N_{\max}^t = 5 \) ($\forall t$) is set for \LaTeXstyle{MATPOWER} tasks, reflecting the increased complexity of the latter. {\color{black}In addition, the scoring was conducted manually by human experts during the evaluation process. }

{\color{black}
\noindent {\color{black}\textbf{Remark 3}}: To further clarify what constitutes the aforementioned irrelevant settings, preliminary examples are provided here for illustration, with full details to be shown in the subsequent case studies. These include: (i) explicitly setting parameters to their default values when not required (e.g., in \LaTeXstyle{DALINE}, setting \texttt{data.baseType} as \texttt{TimeSeriesRand} in Standard Task 2, and \texttt{data.program} as \texttt{acpf} in Standard Task 7, both by \texttt{GPT4o-PR}); and (ii) including unnecessary function calls, such as using \texttt{define\_constants} in \LaTeXstyle{MATPOWER} when no constants are actually used (e.g., Standard Task 27 by \texttt{GPT4o-Sole}). Such settings reflect imperfections in code generation and therefore receive partial credit.

}

\subsection{Evaluation on \LaTeXstyle{Daline}}

The evaluation results on \LaTeXstyle{Daline} are illustrated in Fig.~\ref{fig:violinDaline} and Fig.~\ref{fig:barDaline}. Fig.~\ref{fig:violinDaline} depicts the distribution of scores achieved across attempts for each evaluated scheme, differentiating between complex and standard tasks. Fig.~\ref{fig:barDaline} presents the success rates for each scheme, itemized by ``all tasks combined'', ``complex tasks only'', ``standard tasks only'', as well as for the ``first attempt success rate'' and the ``final attempt success rate''. In the following, these evaluation results are analyzed from multiple perspectives.

\subsubsection{Original Capability vs. Enhanced Capability}

While equipped with environmental interaction and feedback mechanisms, \texttt{GPT4o-Sole} still demonstrates a 0\% success rate for both complex and standard tasks, indicating that GPT4o has not previously encountered \LaTeXstyle{Daline}. Even with a complete knowledge base supported by RAG — either through the standard RAG or OpenAI's official RAG — the resulting schemes, \texttt{GPT4o-SR} and \texttt{CGPT4o-R}, achieve success rates of only 31.37\% and 33.82\% across all tasks, respectively. This suggests that, even with RAG support, the latest language model, GPT4o, still lacks reliable performance in simulations. In contrast, the scheme equipped with the proposed full framework, \texttt{GPT4o-Full}, achieves a success rate of 93.13\% across all tasks — a significant improvement that highlights the effectiveness of the proposed framework.

\subsubsection{Fully Equipped vs. Less Equipped} The high success rate of 93.13\% over all {\color{black}tasks} for \texttt{GPT4o-Full} is due to the cumulative effects of using the complete proposed framework. Comparing other schemes with \texttt{GPT4o-Full} gives an indication of the impact of omitted strategies. For instance, although \texttt{GPT4o-NP} includes most reasoning enhancement strategies, it lacks the triple-based structured option document, resulting in a reduced success rate of 81.37\%. On the other hand, omitting the few-shot CoT for reasoning, as in \texttt{GPT4o-NCS}, lowers success to 65.19\%. When few-shot CoT is employed, but the proposed query planning is omitted, as in \texttt{GPT4o-RSR}, the success rate for complex tasks drops to 66.67\%, particularly due to the deteriorated performance for the complex tasks. Similar comparisons can be drawn across all schemes. However, the goal is not to determine which strategy provides the highest improvement. Instead, {\color{black}this paper aims to} emphasize that high success relies on the combined effect of multiple strategies.

\subsubsection{Complex Tasks vs. Normal Tasks} 
\begin{figure*}[t!]
\centering
\includegraphics[width=1\linewidth]{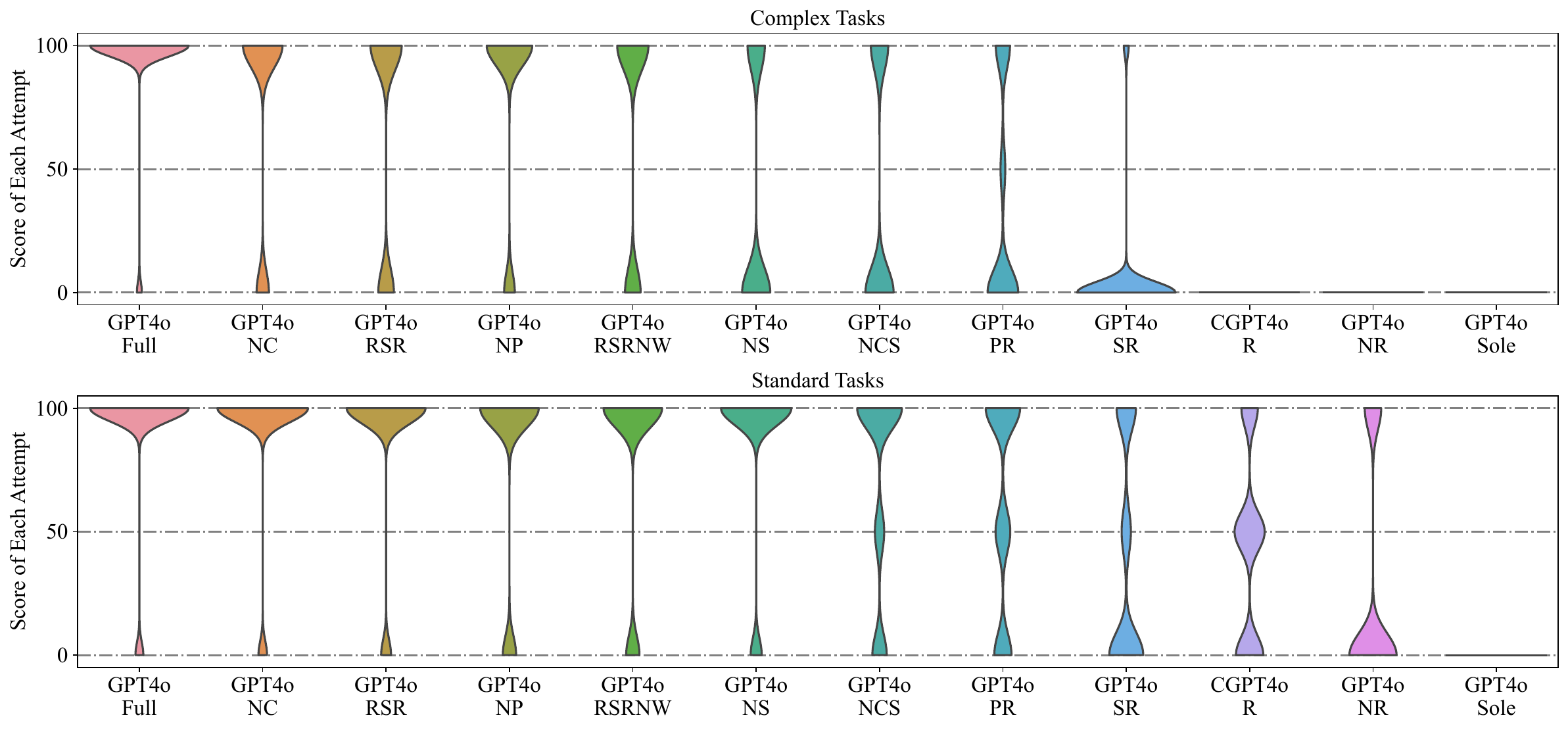} 
\caption{{\color{black}Distribution of scores achieved across attempts for each evaluated scheme. For the definitions of scheme notations (e.g., \texttt{GPT4o-Full}), refer to Table~\ref{tab:strategy}; the notations apply hereafter. (simulation environment: \LaTeXstyle{Daline}; GPT4o version:  gpt-4o-2024-05-13)}.}
\vspace{-4mm}
\label{fig:violinDaline}
\end{figure*}

\begin{figure*}[t!]
\centering
\includegraphics[width=1\linewidth]{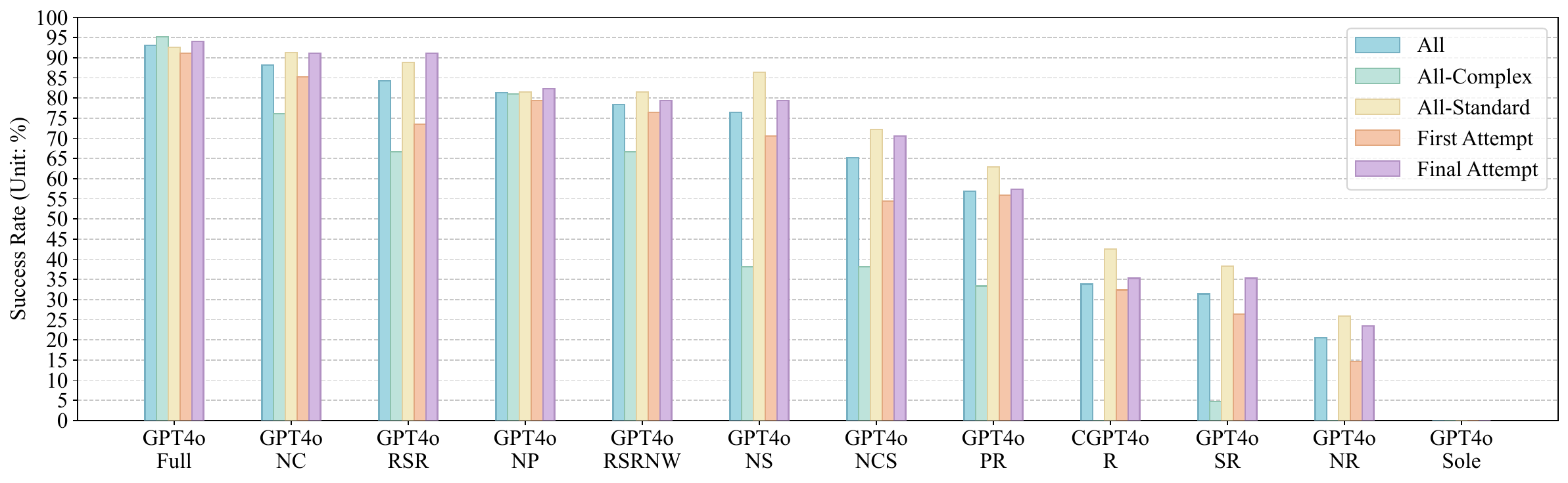} 
\caption{{\color{black}Success rates for each scheme. ``All" refers to the aggregated success rate across all tasks; ``All-Complex" and ``All-Standard" report the success rates calculated exclusively for complex and standard tasks, respectively. ``First Attempt" represents the success rate achieved on the first attempt, before any error correction, and ``Final Attempt" reflects the success rate after all permitted attempts are completed. The same interpretation applies hereafter (simulation environment: \LaTeXstyle{Daline}; GPT4o version:  gpt-4o-2024-05-13)}. }
\vspace{-4mm}
\label{fig:barDaline}
\end{figure*}

In general, more complex tasks (i.e., those with multiple sub-requests) tend to increase the likelihood of errors in LLMs, resulting in lower success rates compared to standard tasks, as observed across most schemes. However, with {\color{black}the} proposed full framework, \texttt{GPT4o-Full}, the performance gap between complex and standard tasks narrows significantly, as shown in both Fig.~\ref{fig:violinDaline} and Fig.~\ref{fig:barDaline}. This suggests that, with enhanced reasoning capabilities and the more effective RAG design, \texttt{GPT4o-Full} effectively identifies and addresses the sub-requests within complex tasks, similar to how it handles standard tasks. This enables LLMs to better manage complex tasks. 

\FloatBarrier

\begin{figure*}[t!]
\centering
\includegraphics[width=1\linewidth]{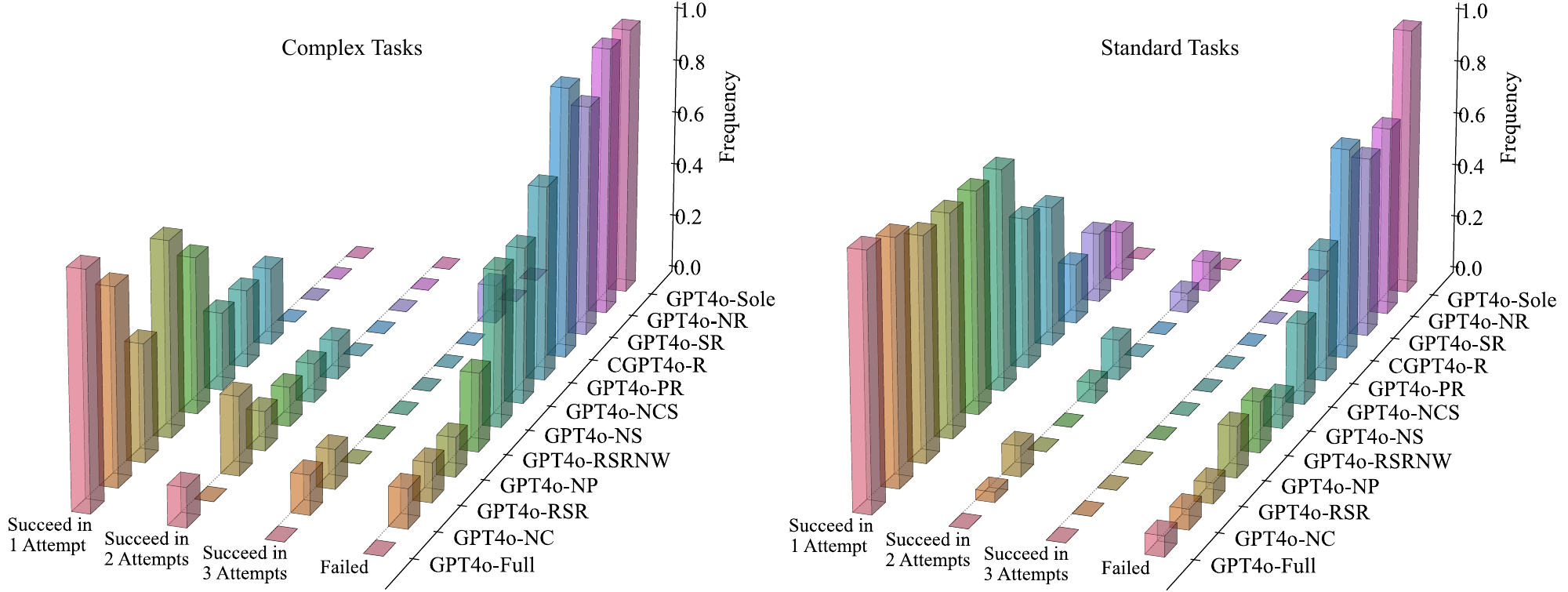} 
\caption{{\color{black}Distribution of feedback loops (attempts) required for successful task completion in complex (left) and standard (right) tasks. (simulation environment: \LaTeXstyle{Daline}; GPT4o version:  gpt-4o-2024-05-13).} }
\vspace{-4mm}
\label{fig:bar3d_daline}
\end{figure*}

\subsubsection{First Attempt vs. Final Attempt} The comparison between the first-attempt and final-attempt success rates demonstrates the effectiveness of environmental interaction and feedback mechanisms. As shown in Fig.~\ref{fig:barDaline}, the final-attempt success rate is always higher than the first-attempt rate, particularly for schemes that incorporate fewer strategies from the proposed framework. These schemes typically have either reduced reasoning capability or limited retrieval information, making environmental interaction and feedback crucial for error correction. However, for \texttt{GPT4o-Full}, the difference between first-attempt and final-attempt success rates is relatively small, as \texttt{GPT4o-Full} often completes the \LaTeXstyle{Daline} simulation task successfully on the first attempt. This further highlights the effectiveness of the proposed framework. One noteworthy point is that the effectiveness of automatic error correction depends partly on the quality of the simulation tool's error-reporting system. Specifically, it matters whether the system provides clear, code-specific error messages. This feature affects the LLM's capability to interpret and resolve issues in the generated code. In the absence of such a feature, as with \texttt{GPT4o-RSRNW}, the success rate drops to 78.43\%, with negligible improvement between the first and final attempts. This underscores that without a well-developed error-reporting system, iterative refinement may yield limited benefit. Although tools like \LaTeXstyle{Daline} and \LaTeXstyle{MATPOWER} include well-developed error reporting (thereby achieving high correction accuracy), for other simulation tools where such systems are less developed, reinforcing them is recommended. In fact, the proposed framework also enables LLMs to detect vulnerabilities within a simulation tool's error-reporting system, particularly when mistakes occur. These errors may not stem from limitations of the LLMs or the framework itself but rather from inherent design issues within the tools. 

{\color{black}
\subsubsection{Distribution of Attempts for Task Completion}

To further examine the feedback mechanism, Fig.~\ref{fig:bar3d_daline} presents the distribution of attempts (feedback loops) required to successfully solve complex and standard tasks. For complex tasks, baseline schemes (e.g., \texttt{GPT4o-Sole}) frequently fail or require multiple attempts, indicating that resolving complicated queries (i.e., tasks with more sub-queries and dependencies) poses significant challenges. In contrast, \texttt{GPT4o-Full} performs markedly better, with most tasks completed within one or two attempts. For standard tasks, all schemes achieve improved results, and failures are infrequent. \texttt{GPT4o-Full} again shows clear advantages, solving the majority of standard tasks in a single attempt. Even less-equipped schemes perform relatively well on standard tasks, suggesting that these tasks impose lower demands on reasoning, retrieval, and error correction, and are thus more manageable for LLMs.

}

\noindent {\color{black}\textbf{Remark 4}}: Enhancing error-reporting systems in simulation tools is beyond this paper's scope. However, for tools with ambiguous error messages, the following is proposed:
	\begin{itemize}[leftmargin=10pt]
		\item \textbf{If internal modifications are feasible}, a layered checking mechanism can be implemented, where an agent evaluates error reports across functional layers, identifies vulnerabilities, and iteratively refines messages.
		\item \textbf{If internal modifications are not feasible}, three external strategies may be employed: (i) For LLMs with long-context capabilities, sending the entire underlying code for analysis is possible but costly. (ii) A more efficient alternative is to batch extract relevant code segments—from surface-layer to deeper-layer functions—and submit them alongside error messages to the agents for analysis. (iii) An active trial-and-error mechanism based on agents can automatically test code with intentional bugs, mapping observed errors to ambiguous messages, in order to flag problematic outputs and facilitate learning.
	\end{itemize}
These strategies can be seamlessly integrated into {\color{black}the proposed} framework. Internal modifications require no changes to the framework since they occur within the simulation tool. Notably, the layered checking mechanism has already been implemented in \LaTeXstyle{Daline}, proving highly effective not only for the proposed framework but also as a robust enhancement for \LaTeXstyle{Daline} itself. For external strategies, the only adjustment needed is to extend the error report with relevant code extraction or error-message mapping. Future research will focus on leveraging the LLM-based trial-and-error mechanism to systematically map ambiguous error messages, reducing diagnosis overhead and improving robustness.

\begin{figure*}[t]
    \centering
    \includegraphics[width=1\linewidth]{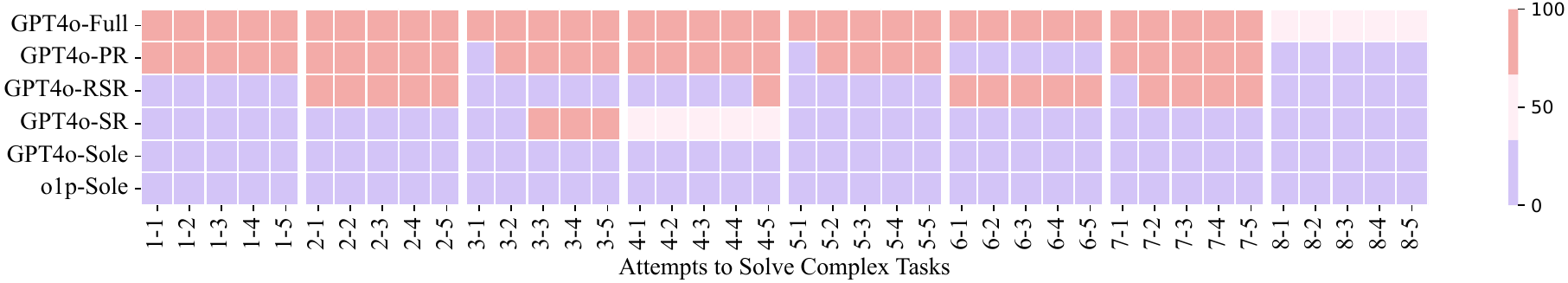} 
    \caption{Scores achieved by each evaluated scheme in individual attempts when handling complex tasks. The automatic error correction mechanism aids schemes in correcting their behavior when encountering errors in initial attempts  {\color{black}(simulation environment: \LaTeXstyle{MATPOWER}; GPT4o version:  gpt-4o-2024-05-13)}. }
    \vspace{-4mm}
    \label{fig:heatmap}
    \end{figure*}
    
    \begin{figure*}[t]
    \centering
    \includegraphics[width=1\linewidth]{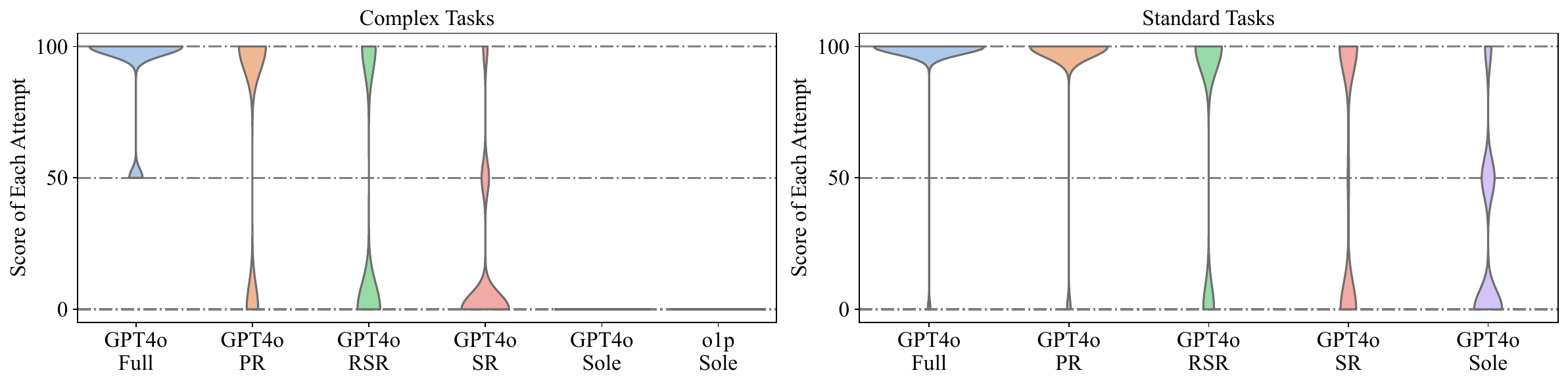} 
    \caption{Distribution of scores achieved across attempts for each evaluated scheme, separated by complex and standard tasks  {\color{black}(simulation environment: \LaTeXstyle{MATPOWER}; GPT4o version: gpt-4o-2024-05-13)}. }
    \vspace{-4mm}
    \label{fig:violinMatpower}
    \end{figure*}
    
    \begin{figure}[t]
    \centering
    \includegraphics[width=1\linewidth]{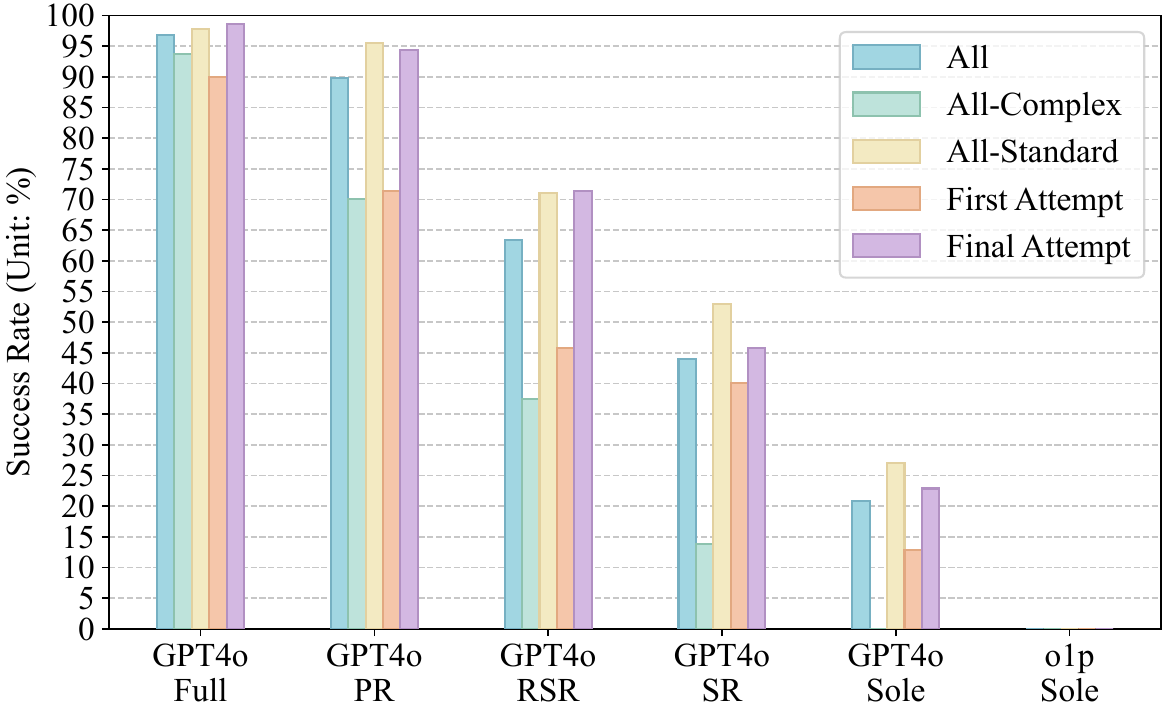} 
    \caption{Success rates for each scheme, broken down by all tasks combined, complex tasks only, standard tasks only, as well as for the first attempt success rate and the final attempt success rate {\color{black}(simulation environment: \LaTeXstyle{MATPOWER}; GPT4o version: gpt-4o-2024-05-13; \texttt{o1p-Sole} is only tested by complex tasks)}.}
    \vspace{-4mm}
    \label{fig:barMatpower}
    \end{figure}

\subsection{Evaluation on \LaTeXstyle{MATPOWER}}

The evaluation results on \texttt{MATPOWER} are illustrated in Figs. \ref{fig:heatmap}, \ref{fig:violinMatpower}, and \ref{fig:barMatpower}. Specifically, Fig.~\ref{fig:heatmap} presents the scores achieved by each evaluated scheme in individual attempts when managing complex tasks. Fig.~\ref{fig:violinMatpower} shows the distribution of scores across attempts for each scheme, and Fig.~\ref{fig:barMatpower} depicts the success rates of each scheme. {\color{black} The outcomes observed here align closely with the results on \LaTeXstyle{Daline}. This alignment allows us to focus primarily on comparative analyses across schemes in the subsequent discussion. }

\subsubsection{Original Capability vs. Enhanced Capability}

Despite \LaTeXstyle{MATPOWER} being a widely-used and well-documented tool with extensive resources available online, the latest high-performance LLMs, such as \texttt{GPT4o} and \texttt{o1-preview} (renowned for the reasoning capability),  struggle to perform simulations reliably. For instance, both \texttt{GPT4o-Sole} and \texttt{o1p-Sole} show a 0\% success rate on complex tasks, and \texttt{GPT4o-Sole} achieves only 27.77\% success on standard tasks. Even with RAG and the whole knowledge base, \texttt{GPT4o-SR} reaches a success rate of only 13.75\% for complex tasks and 52.96\% for standard tasks. In contrast, the fully equipped framework, \texttt{GPT4o-Full}, achieves a remarkable 96.85\% success rate across all tasks, with a breakdown of 93.75\% on complex tasks and 97.77\% on standard tasks.

\subsubsection{Fully Equipped vs. Less Equipped}

Consistent with the findings on \LaTeXstyle{Daline}, the results on \LaTeXstyle{MATPOWER} indicate that high success rates depend on the synergistic effect of multiple strategies. For example, excluding the enhanced reasoning module, as in \texttt{GPT4o-PR}, results in an overall success rate decrease to 89.71\%, with complex tasks dropping further to 70.00\%. Similarly, omitting the proposed query planning strategy, as in \texttt{GPT4o-RSR}, reduces the overall success rate to 63.42\% and complex tasks to 37.50\%. These outcomes are substantially lower than those achieved by \texttt{GPT4o-Full}, which maintains a 93.75\% success rate on complex tasks and 97.77\% on standard tasks, demonstrating the critical role of each component within the proposed framework.

\begin{figure}[!t] 
    \centering
    \includegraphics[width=1\linewidth]{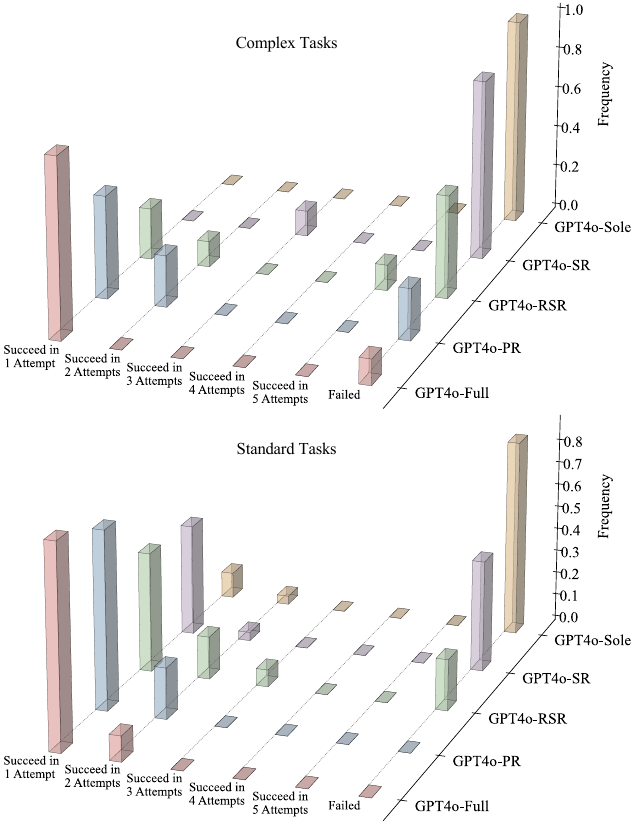} 
    \caption{{\color{black}Distribution of feedback loops (attempts) required for successful task completion in complex (upper) and standard (lower) tasks. (simulation environment: \LaTeXstyle{MATPOWER}; GPT4o version: gpt-4o-2024-05-13; \texttt{o1p-Sole} is only tested by complex tasks)}.}
    \vspace{-4mm}
    \label{fig:bar3d_matpower}
    \end{figure}

{\color{black}
\subsubsection{Distribution of Attempts for Task Completion}

Compared to Fig.~\ref{fig:bar3d_daline}, a similar pattern of the attempt distribution is observed under the \LaTeXstyle{MATPOWER} simulation environment, as shown in Fig.~\ref{fig:bar3d_matpower}. \texttt{GPT4o-Full} still maintains a clear advantage by completing the majority of tasks in the first or second attempt. These results further confirm the effectiveness and generalizability of the proposed framework across different simulation environments.

}

\subsection{Comparison with Supervised Fine-tuning}
To further verify the proposed framework's performance, it was compared with supervised fine-tuning (SFT)\footnote{{\color{black}The OpenAI's SFT approach was used in this paper; see \href{https://platform.openai.com/docs/guides/fine-tuning}{[\underline{here}]} for details.}}, a well‐established approach known to enhance LLM performance on specific tasks—including unseen tasks such as translation and natural language inference \cite{wei2021finetuned}. In order to precisely demonstrate the improvements introduced by SFT, \texttt{GPT4o-Sole} has been employed as the foundation for SFT training\footnote{{\color{black}Since gpt-4o-2024-08-06 is the only version available for SFT, all GPT4o-related schemes in Section-VD use this version for consistency.}}, denoted by \texttt{GPT4o-Sole-SFT} in Table~\ref{tab:strategy}. 

\begin{figure}[t]
    \centering
    \includegraphics[width=1\linewidth]{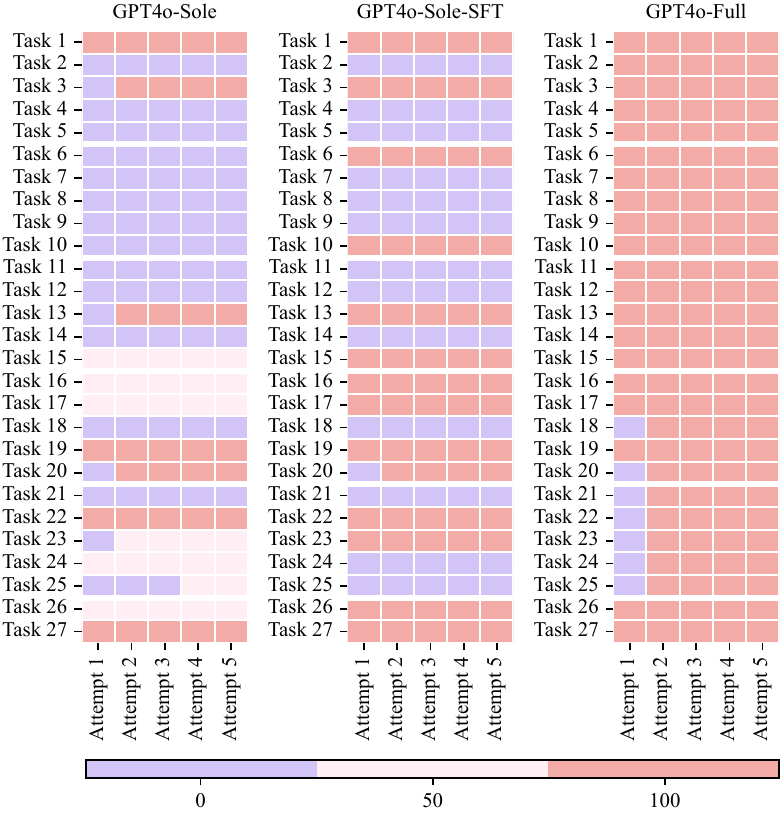} 
    \caption{{\color{black}Scores achieved by each evaluated scheme in individual attempts when handling standard tasks (simulation environment: \LaTeXstyle{MATPOWER}; GPT4o version: gpt-4o-2024-08-06).}}
    \label{fig:heat0806}
    \end{figure}

\begin{figure}[t]
    \centering
    \includegraphics[width=1\linewidth]{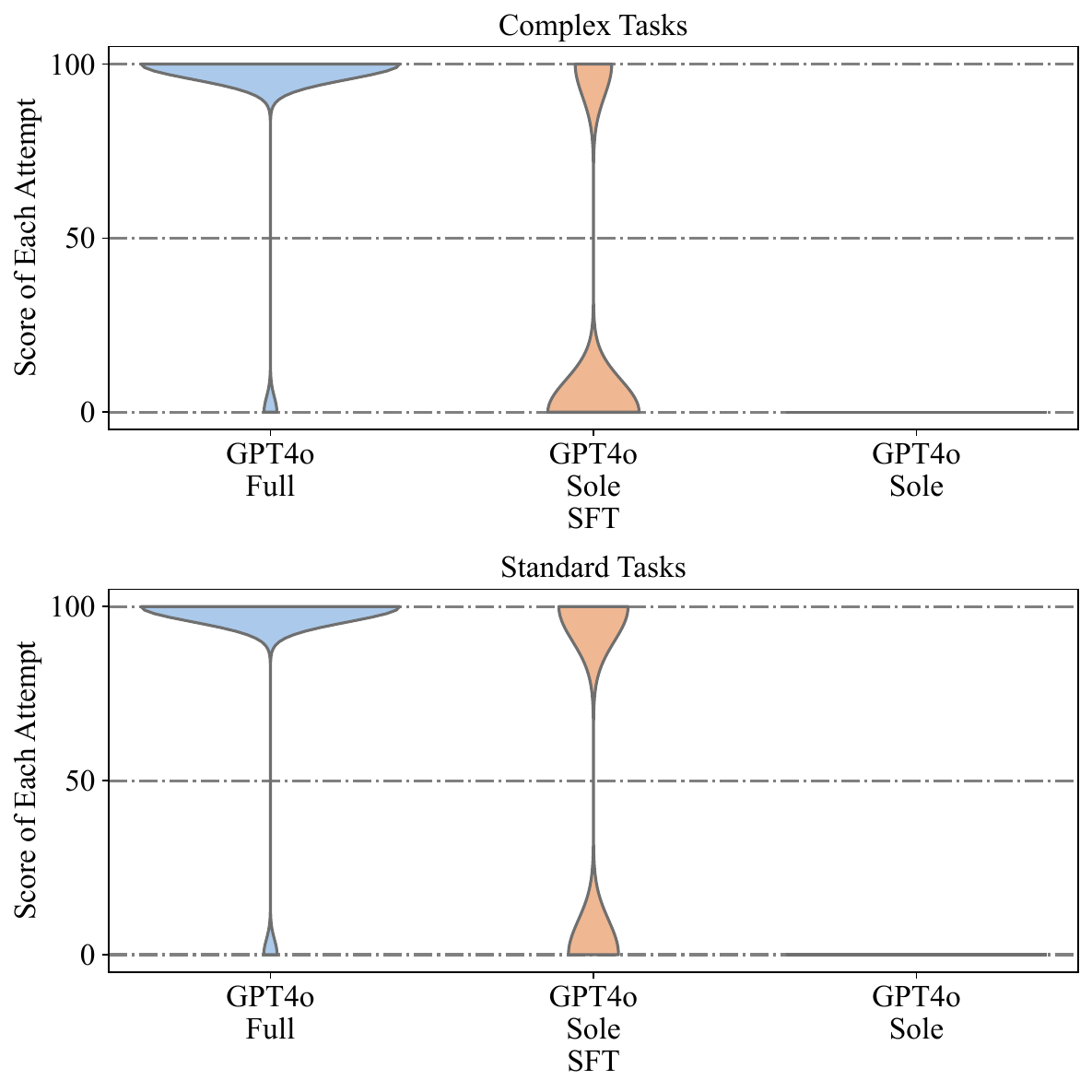} 
    \caption{{\color{black}Distribution of scores achieved across attempts for each evaluated scheme (simulation environment: \LaTeXstyle{Daline}; GPT4o version: gpt-4o-2024-08-06).}}
    \vspace{-4mm}
    \label{fig:violin_sft}
    \end{figure}

The initial comparison was conducted on \LaTeXstyle{MATPOWER} tasks. For this evaluation, 50 random tasks (following the recommendation from OpenAI) were generated with preferred coding responses, producing a supervised training dataset of 24,249 tokens. The model was fine-tuned for 3 epochs, with a batch size of 1 and a learning rate multiplier of 2 (all automatically configured). Fig.~\ref{fig:heat0806} illustrates the performance of \texttt{GPT4o-Full}, \texttt{GPT4o-Sole}, and \texttt{GPT4o-Sole-SFT}, based on the scores achieved in individual attempts on standard tasks. Evidently, SFT enhances the performance of \texttt{GPT4o-Sole}: \texttt{GPT4o-Sole-SFT} consistently attains more perfect scores (100 points) than \texttt{GPT4o-Sole}, with success rates of 51.11\% versus 35.18\%, respectively. However, \texttt{GPT4o-Full} significantly outperforms \texttt{GPT4o-Sole-SFT}, achieving 100 points for all tasks starting from the second attempt. 

\begin{table}[t]
    {\color{black}
\centering
\footnotesize
\caption{{\color{black}Success rates for each scheme, broken down by all tasks combined, complex tasks only, standard tasks only, as well as for the first attempt success rate and the final attempt success rate (simulation environment: \LaTeXstyle{Daline}; GPT4o version: gpt-4o-2024-08-06)}}
\vspace{-2pt}
\renewcommand{\arraystretch}{1.2} 
\setlength{\tabcolsep}{8pt}
\label{tab:score_sft}
\begin{tabular}{cccc}
\hline
\textbf{Schemes} & \textbf{GPT4o-Sole} & \textbf{GPT4o-Sole-SFT} & \textbf{GPT4o-Full} \\ \hline
All-Tasks & 0.000\% & 51.961\% & 95.098\% \\
All-Complex & 0.000\% & 28.571\% & 95.238\%  \\
All-Standard & 0.000\% & 58.025\% & 95.062\% \\
First Attempt & 0.000\% & 50.000\% & 91.176\% \\
Final Attempt & 0.000\% & 52.941\% & 97.059\% \\
\hline
\end{tabular}
    }
\end{table}

  \begin{figure*}[!b]
\centering
\includegraphics[width=1\linewidth]{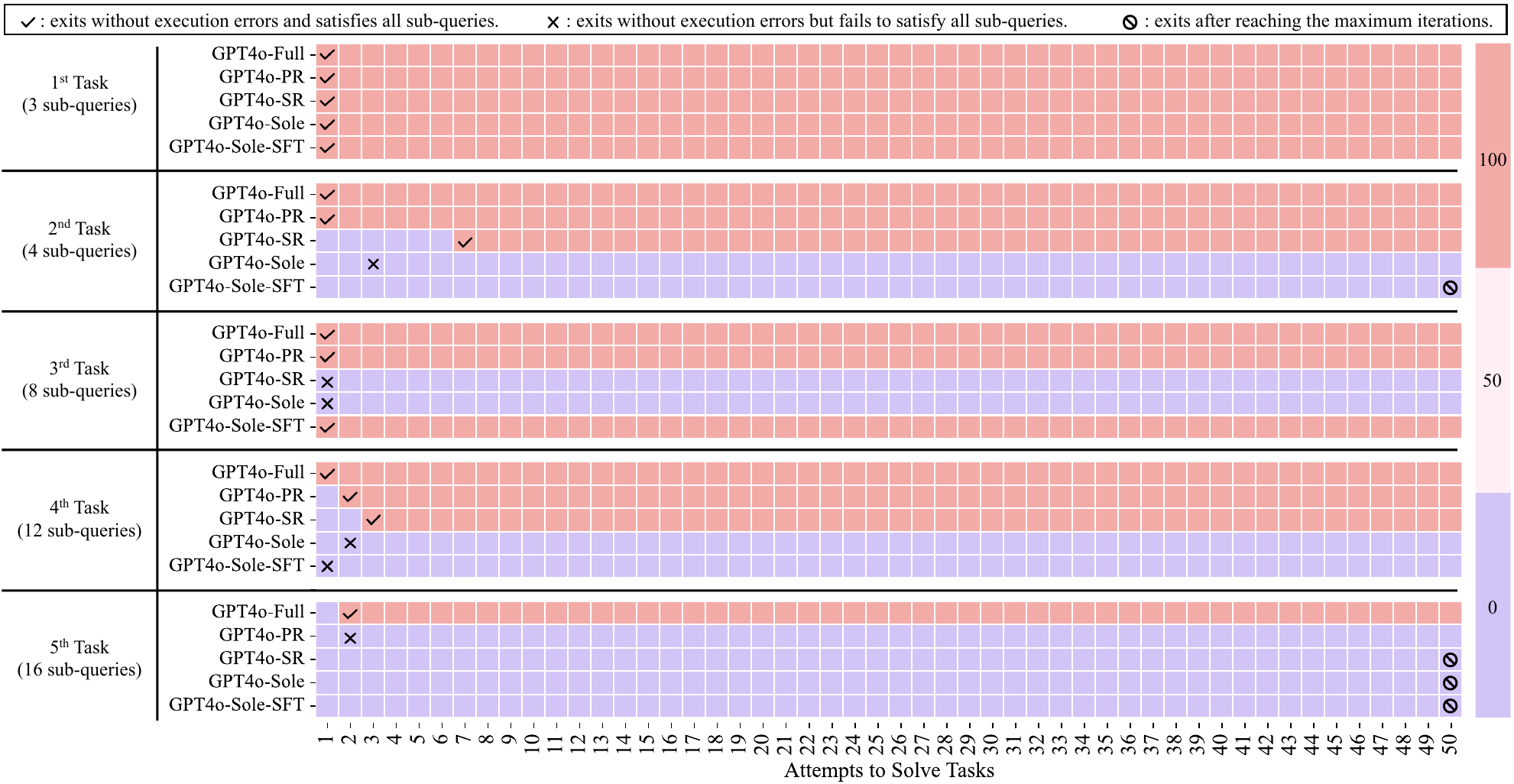} 
\caption{{\color{black}Scores achieved by each evaluated scheme in individual attempts when handling tasks with different difficulty levels; the bar on the right represents the score scale, where 100, 50, and 0 indicate different score values (simulation environment: \LaTeXstyle{MATPOWER}; GPT4o version: gpt-4o-2024-08-06)}. }
\vspace{-4mm}
\label{fig:heat_distribution}
\end{figure*}
 
The underperformance of \texttt{GPT4o-Sole-SFT} could stem from the representativeness of the training datasets—despite significant human effort in generating them, these datasets may not and cannot capture every nuanced detail of \LaTeXstyle{MATPOWER}. To rigorously eliminate the possibility that the limited dataset is responsible for the performance gap, an additional evaluation was conducted using the \LaTeXstyle{Daline} simulation environment. In this experiment, the full testing dataset (i.e., the \LaTeXstyle{Daline} simulation tasks used for evaluation) was used as the SFT training dataset. Specifically, all 34 testing simulation tasks, augmented with 16 additional tasks, were employed for fine-tuning, yielding a training dataset of 32,346 tokens. \texttt{GPT4o-Sole-SFT} was fine-tuned for 3 epochs, with a batch size of 1 and a learning rate multiplier of 2 (again, all automatically configured). The score distributions and specific success rates are presented in Fig.~\ref{fig:violin_sft} and Table~\ref{tab:score_sft}. This evaluation shows that \texttt{GPT4o-Sole-SFT} improves the success rate of \texttt{GPT4o-Sole} from 0\% to 51.96\% overall, with 28.57\% for complex tasks and 58.03\% for standard tasks. Yet, even with the entire testing dataset used for {\color{black}fine-tuning}, \texttt{GPT4o-Sole-SFT} still fails to reach high accuracy.

The underperformance of \texttt{GPT4o-Sole-SFT} can be attributed to the inherent limitations of SFT for LLMs: SFT can capture coarse‐grained information such as style, tone, format, or other qualitative aspects, but SFT cannot memorize every nuance of the training data. The reason is twofold: (1) SFT is a lossy compression process, which inevitably discards finer details, and (2) SFT only adjusts a small subset of the model's parameters, with the rest remaining focused on generating generalizable patterns --- as a result, SFT cannot achieve 100\% retention of task-specific details, especially for tasks like coding, which require high-precision parameters and functions. In contrast, \texttt{GPT4o-Full} leverages an enhanced RAG module, allowing the model to dynamically access and retrieve precise information from external documents. This retrieval mechanism effectively mitigates the limitations of parameter-based memorization, enabling \texttt{GPT4o-Full} to achieve over 95\% accuracy across both complex and standard tasks.

{\color{black}

\subsection{Evaluation under Increased Attempt Budget}

To further analyze how tasks evolve toward either successful completion or termination by the stopping criterion, Fig.~\ref{fig:heat_distribution} presents the scores achieved in individual attempts with an extended attempt budget of up to 50. Meanwhile, task difficulty is explicitly characterized by the number of sub-queries within each task. This analysis offers a more comprehensive view of how different schemes perform under varying task complexities, especially when granted extensive opportunities for error correction.

The results in Fig.~\ref{fig:heat_distribution} reveal a clear distinction between schemes and task difficulty levels. Baseline methods struggle significantly with complex tasks --- they often either reach the maximum iteration limit (i.e., the stopping criterion) or terminate without execution errors but still fail to satisfy all sub-queries. Typical failure reasons include: (i) invalid option names, which prevent proper execution and remain unresolved even after 50 feedback loops (e.g., \texttt{GPT4o-Sole-SFT} in the 2nd and 5th tasks, \texttt{GPT4o-Sole} in the 5th task, and \texttt{GPT4o-SR} in the 5th task); (ii) correct option names with invalid parameter values produce incorrect but executable code (e.g., \texttt{GPT4o-Sole} in the 2nd to 4th tasks and \texttt{GPT4o-SR} in the 3rd task); (iii) missing critical option settings, which result in early termination without errors but produce incorrect outputs (e.g., \texttt{GPT4o-PR} in the 5th task).

These observations confirm that repeated error-feedback loops alone are insufficient for convergence on challenging tasks without strong reasoning and retrieval capabilities. In contrast, \texttt{GPT4o-Full} exhibits robust performance across all difficulty levels, completing all tasks --- including those with numerous sub-queries --- within only a few attempts. Notably, \texttt{GPT4o-Full} only failed at the first attempt for the 5th task, where it misused an option name. This error, however, was successfully corrected in the second attempt.

}

\subsection{Cost Analysis}

The cost analysis of \texttt{GPT4o-Full} for executing simulation tasks in \LaTeXstyle{Daline} and \LaTeXstyle{MATPOWER} is presented in Table \ref{tab:cost}, with average values across all tasks shown. The reported time covers the entire process, including retrieval, reasoning, code generation, simulation execution, result aggregation, and, where necessary, code correction. Remarkably, \texttt{GPT4o-Full} completes each task in approximately half a minute. Additionally, the token expense per task is roughly 0.014 USD. 

It is noteworthy that, aside from parallel retrieval, no specialized acceleration techniques were employed in this framework. Thus, despite its already satisfactory performance, there is considerable potential for speed enhancements. Even in the current state, without any specific acceleration strategies, \texttt{GPT4o-Full} can execute approximately 120 simulation tasks per hour in \LaTeXstyle{Daline} and \LaTeXstyle{MATPOWER}, with a total token cost of around 1.68 USD. Given this high efficiency and cost-effectiveness, the proposed framework presents a promising pathway to improve researcher productivity.

It must be emphasized, however, that the reported execution time and token cost serve only as approximate indicators, given the inherent variability introduced by factors such as retrieval, reasoning, code execution, result aggregation, and iterative error correction. This analysis aims to provide readers with a general sense of the efficiency of automated simulations within the scope of the specific simulation tasks, rather than to establish precise quantitative relationships.

\begin{table}[t]
    \centering
    {\color{black}
    \footnotesize
    \begin{threeparttable}
        \caption{{\color{black} Average cost analysis of \texttt{GPT4o-Full} per task$^*$}}
        \vspace{-2pt}
        \renewcommand{\arraystretch}{1.5} 
        \setlength{\tabcolsep}{1.2pt}
        \label{tab:cost}
        \begin{tabular}{ccccc}
            \toprule
            \textbf{Environment} & \textbf{Time (sec.)} & \textbf{Input Token$^\dagger$} & \textbf{Output Token} & \textbf{Expense (USD) $^\ddagger$} \\ \midrule
            \LaTeXstyle{Daline}   & 29.446               & 7882.294             & 168.353               & 0.014                \\
            \LaTeXstyle{MATPOWER} & 32.703               & 5338.514             & 274.371               & 0.013                \\ \bottomrule
        \end{tabular}
        \begin{tablenotes}
            \footnotesize
            \item[$\dagger$] A token is a fundamental unit of text, typically representing fragments of words. Tokens quantify both the input (text submitted to the LLM via API) and the output (text or code generated by the LLM via API).
            \item[$\ddagger$] Expense refers to the monetary cost incurred for processing these tokens, as charged by the API provider (OpenAI in this study).
            \item[$*$] The API provider automatically computes the total tokens and associated expenses. The average values here are calculated by dividing the total tokens (and corresponding expenses) used across all simulation tasks with \texttt{GPT4o-Full} by the number of tasks.
        \end{tablenotes}
    \end{threeparttable}}
\end{table}

\section{Conclusion}

This paper addresses the research gap in enhancing LLMs for power system simulations by proposing a feedback-driven, multi-agent framework. It represents the first systematic approach to significantly improve LLMs' simulation capabilities across both familiar and new tools. Validated on 69 diverse simulation tasks from \LaTeXstyle{Daline} and \LaTeXstyle{MATPOWER}, the proposed framework achieved substantial performance improvements, with success rates of 93.13\% and 96.85\%, respectively. {\color{black}It} far surpasses those of baseline schemes, including the latest LLM, o1-preview, {\color{black}and SFT for LLMs}. Key findings include: (i) The original simulation capability of LLMs is limited, as evidenced by \texttt{GPT4o} and \texttt{o1-preview} achieving success rates no higher than 27.77\%. (ii) Even with the standard RAG module and a comprehensive knowledge base, LLMs achieve overall success rates below 45\%, highlighting the need for a more comprehensive approach. (iii) Our framework's high success rate stems from a synergistic integration of enhanced RAG, enhanced reasoning, as well as environmental acting and feedback mechanisms. Removing any of these elements results in a significant performance decline. (iv) While SFT may be able to capture coarse-grained attributes such as style, tone, and format, it struggles to retain the full detail necessary for power system simulations. Even when the entire test dataset is included in the training set, SFT achieves a success rate of less than 60\%. This limitation stems from its inherently lossy compression mechanism and constrained parameter update capacity, which prevent it from encoding fine-grained details with high fidelity. (iv) The proposed framework enables LLMs to execute tasks efficiently, with each task completed in approximately 30 seconds at a token cost of only 0.014 USD (for the simulation tasks used in this paper), offering a scalable, cost-effective solution that enhances {\color{black}the productivity} of human scientists in power systems.

{\color{black}
Overall, the effectiveness of this work relies on the coordinated integration of general-purpose LLM capabilities (such as natural language understanding and code generation) with carefully designed domain-specific strategies tailored to power system simulations. These strategies are essential for addressing the unique challenges posed by simulation tasks. While this work focuses on simulations, the proposed framework is inherently designed to enable LLMs to adapt to unfamiliar domains and is readily extensible to broader power system research. Furthermore, the proposed simulation-enhanced LLM system may serve as a key component in future human-machine collaborative research, supporting not only simulation execution but also the validation and refinement of research ideas.} 

However, several critical future challenges still remain. \textbf{First}, while the proposed framework demonstrates significant improvements over existing schemes and supervised fine-tuning, achieving a perfect 100\% success rate for LLM-based simulations remains an open challenge. For instance, the proposed framework sometimes fails to detect ``non-execution-bug'' failures, resulting in unaware, inaccurate outputs that cannot be automatically corrected. A potential avenue for future work is to explore the use of a RAG-supported, fine-tuned LLM-based code-checking agent, which would verify executed simulation code and identify hidden errors that the error-reporting mechanism may overlook. \textbf{Second}, when simulation requests are inherently ambiguous or underspecified, accurately determining the user's intent becomes difficult for any automated method, not just the framework presented in this paper. This ambiguity primarily arises from the initial lack of detail in the user's input, rather than any intrinsic limitation of natural language coding. To address this, a promising future direction is to integrate an interactive dialogue stage into the framework. In this approach, an autonomous agent would evaluate the specificity of the initial request and, if necessary, prompt the user with targeted clarifying questions. This human-in-the-loop strategy could help refine the requirements, leading to more precise retrieval and improved code generation.


\section*{Acknowledgement}
We would like to acknowledge the assistance of ChatGPT-4o \cite{openai2024chatgpt4o} for language polishing of this paper.

\bibliographystyle{IEEEtran}
\bibliography{IEEEabrv,paper}
\end{document}